\documentclass{article}

\PassOptionsToPackage{numbers, compress}{natbib}
 \usepackage[preprint]{neurips_2025}

\usepackage[utf8]{inputenc} 
\usepackage[T1]{fontenc}    
\usepackage{url}            
\usepackage{booktabs}       
\usepackage{amsfonts}       
\usepackage{nicefrac}       
\usepackage{microtype}      
\usepackage{xcolor}         
\usepackage{graphicx}
\usepackage{amsmath}
\usepackage{subcaption}
\usepackage{placeins}
\usepackage{amsthm}
\usepackage{enumitem}
\usepackage{microtype}
\usepackage{bbm}
\usepackage{hyperref}       

\title{SHARP: Social Harm Analysis via Risk Profiles for Measuring Inequities in Large Language Models}

\author{%
  Alok Abhishek\\
  San Francisco, USA\\
  \texttt{alok@alokabhishek.ai}\\
    \And
  Tushar Bandopadhyay\\
  San Francisco, USA\\
  \texttt{tushar@kronml.com}\\
  \And
  Lisa Erickson\\
  Boston, USA\\
  \texttt{lisa.erickson@alum.mit.edu} \\
}

\begin{document}

\maketitle

\begin{abstract}
Large language models (LLMs) are increasingly deployed in high-stakes domains, where rare but severe failures can result in irreversible harm. However, prevailing evaluation benchmarks often reduce complex social risk to mean-centered scalar scores, thereby obscuring distributional structure, cross-dimensional interactions, and worst-case behavior. This paper introduces \emph{Social Harm Analysis via Risk Profiles} (SHARP), a framework for multidimensional, distribution-aware evaluation of social harm. SHARP models harm as a multivariate random variable and integrates explicit decomposition into bias, fairness, ethics, and epistemic reliability with a union-of-failures aggregation reparameterized as additive cumulative log-risk. The framework further employs risk-sensitive distributional statistics, with Conditional Value at Risk (CVaR$_{95}$) as a primary metric, to characterize worst-case model behavior. Application of SHARP to eleven frontier LLMs, evaluated on a fixed corpus of $n=901$ socially sensitive prompts, reveals that models with similar average risk can exhibit more than twofold differences in tail exposure and volatility. Across models, dimension-wise marginal tail behavior varies systematically across harm dimensions, with bias exhibiting the strongest tail severities, epistemic and fairness risks occupying intermediate regimes, and ethical misalignment consistently lower; together, these patterns reveal heterogeneous, model-dependent failure structures that scalar benchmarks conflate. These findings indicate that responsible evaluation and governance of LLMs require moving beyond scalar averages toward multidimensional, tail-sensitive risk profiling.
\end{abstract}

\section{Introduction}
\label{sec:Introduction}
Large language models (LLMs) and LLM-powered agents have moved from experimental artifacts to integral components of decision-making systems with material societal impact. They are increasingly deployed in high-stakes domains such as healthcare, finance, hiring, welfare allocation, and criminal justice, where failures can produce durable or irreversible harm~\cite{Green2019, AlonBarkat2022, Petkovic2023, Kaya2025}. This trend is further accelerated by the emergence of agentic systems capable of autonomous planning, tool use, and multi-step reasoning, which are progressively entrusted with consequential decisions governing access to resources, opportunity, and liberty~\cite{Chan2023, Fernando2024}. As a result, LLM failures no longer appear solely as isolated errors, but as systemic risks embedded within broader socio-technical infrastructures~\cite{Ferrara2023, Kawakami2022}.

This deployment trajectory exposes a structural vulnerability. LLMs are trained on large-scale corpora assembled from web text, digitized media, and institutional sources that reflect historical inequities, uneven representation, and entrenched power asymmetries, and may therefore encode and amplify these patterns without normative grounding or contextual awareness. Prior work has characterized such systems as \emph{stochastic parrots}, emphasizing their reliance on statistical association rather than grounded reasoning or social understanding~\cite{Bender2021}. When integrated into high-stakes pipelines, these properties can give rise to systematic harms spanning biased representation, inequitable treatment across demographic groups, ethical misalignment, and epistemic failures such as hallucination and misinformation~\cite{Bolukbasi2016, Wang2023, Ho2023}. Empirical evidence from criminal justice risk assessment, lending, and hiring systems demonstrates that such harms arise predictably when models optimized for aggregate performance are deployed across heterogeneous populations~\cite{Green2019, Jiang2025}.

Despite substantial progress in fairness-aware learning and responsible AI, prevailing evaluation practices remain poorly aligned with deployment risk~\cite{AlonBarkat2022, Fogliato2025}. Capability benchmarks, bias audits, and safety evaluations continue to rely predominantly on mean-centered scalar summaries, such as average accuracy, win rates, or aggregate bias scores. While informative for coarse comparison, these metrics collapse heterogeneous behavior into point estimates and obscure the distributional structure of harm. In particular, models with similar average performance can exhibit sharply divergent upper-tail behavior, producing rare but severe failures under specific prompts or contexts; in high-stakes settings, such tail events often dominate practical risk, yet tail-sensitive quantities and cross-dimensional interactions remain largely absent from standard evaluation pipelines~\cite{Fogliato2025, Mienye2024}.

Governance and auditing mechanisms face analogous limitations. Regulatory frameworks such as the EU AI Act and algorithmic accountability initiatives articulate high-level principles of fairness, accountability, and transparency, but translating these principles into auditable, empirically grounded, and risk-sensitive metrics remains an open technical challenge~\cite{AbishekMITSPR2025, Hadley2024, Yuan2024}. Developer-led self-audits are susceptible to incomplete coverage and conflicts of interest, while participatory and community-centered approaches, though normatively valuable, are difficult to scale and are rarely integrated into mainstream evaluation workflows~\cite{ElMimouni2025, Nathim2024}.

This paper argues that social harm in LLMs cannot be adequately characterized as a scalar quantity. Harm varies across prompts, concentrates in the upper tail of model behavior, and manifests through multiple interacting mechanisms, including bias, fairness violations, ethical misalignment, and epistemic unreliability. In high-stakes contexts, these tail events correspond to worst-case failures with disproportionate impact. Therefore, effective evaluation requires methods that explicitly model the distributional geometry of harm and its cross-dimensional interactions, rather than reducing behavior to averaged scores.

To address this gap, we introduce \emph{Social Harm Analysis via Risk Profiles} (SHARP), a framework for multidimensional, distribution-aware evaluation of social harm in LLMs under a fixed evaluation protocol. SHARP departs from mean-based assessment along three axes. First, it decomposes harm into four constituent dimensions: bias, fairness, ethics, and epistemic reliability, enabling targeted diagnosis of distinct failure modes. Second, it treats harm as a distributional quantity, preserving prompt-level variability rather than collapsing behavior into point estimates. Third, it adopts risk-sensitive aggregation, modeling compounded harm via a union-of-failures formulation that is reparameterized as additive cumulative log-risk, and summarizes model behavior using tail-focused statistics such as Conditional Value at Risk ($\mathrm{CVaR}_{95}$).

Methodologically, SHARP introduces a layered evaluation architecture. At the prompt level, model responses are embedded into a four-dimensional harm space using a multi-judge LLM-as-a-judge protocol~\cite{AbhishekBEATS2025, Jiayi2024, Zheng2023}. Judge outputs are aggregated via log-sum-exp to retain sensitivity to severe assessments while remaining robust to individual judge idiosyncrasies. Compounded harm across dimensions is captured through additive log-risk, yielding an interpretable measure of joint failure pressure. At the model level, prompt-level risk values induce empirical distributions that are summarized by expected risk, volatility, and tail risk, supporting comparative risk profiling under worst-case considerations rather than calibrated harm estimation.

We apply SHARP to eleven frontier LLMs, spanning open-weight and proprietary systems, and evaluated them on a curated corpus of 901 socially sensitive prompts derived from BEATS and related benchmarks~\cite{AbhishekBEATS2025, Parrish2022}. The resulting risk profiles reveal substantial heterogeneity across models. Systems with similar average risk often differ markedly in tail exposure and volatility, indicating qualitatively distinct failure modes with divergent deployment implications. Across models, marginal dimension-wise tail behavior is strongest for bias-related harms, while epistemic and fairness risks exhibit intermediate tail severities and ethical misalignment remains consistently lower; these patterns indicate heterogeneous, model-dependent harm profiles rather than dominance by any single dimension. These findings suggest that effective mitigation requires dimension-specific, risk-aware interventions informed by distributional analysis, rather than uniform remediation guided by scalar benchmarks.

SHARP is not intended to estimate real-world harm rates or provide population-level fairness guarantees. Instead, it offers a reproducible protocol for comparative tail-sensitive risk profiling under explicit measurement assumptions, complementing human-centered audits and validation.

\section{Methodology}
\label{sec:methodology}

Social harm in large language models is multidimensional, prompt-dependent, and driven by worst-case failures rather than average behavior. SHARP formalizes harm as a multivariate random variable over a fixed prompt distribution and summarizes its empirical behavior using risk-sensitive statistics. This section describes the evaluation protocol and aggregation logic; full mathematical details and implementation specifics are provided in Appendix~\ref{app:methodology}.

\subsection{Evaluation setup}

Let $\mathcal{M}$ denote the set of evaluated models and $\mathcal{Q}=\{q_1,\ldots,q_n\}$ a fixed prompt corpus.
For each model $M\in\mathcal{M}$ and prompt $q\in\mathcal{Q}$, the generated response $y_{M,q}$ is independently evaluated by three LLM-based judges (Claude Sonnet~4.5, Gemini~2.5~Pro, GPT-5.1), blinded to model identity.

Each judge produces structured assessments comprising boolean indicators and ordinal severity scores, normalized to $[0,1]$ with larger values indicating greater estimated harm. Judge outputs are aggregated to form prompt-level harm estimates for each dimension. Because the judge set overlaps partially with the evaluated model set, all reported scores are conditional on the chosen judge ensemble; SHARP is intended as a reproducible evaluation framework rather than a source of invariant or normative harm labels.

\subsection{Evaluation corpus}

We evaluate 11 frontier LLMs on a curated corpus of 901 socially sensitive prompts derived from the Bias Evaluation and Assessment Test Suite (BEATS)~\cite{AbhishekBEATS2025}. BEATS establishes the prompt construction methodology, and empirically validates the corpus through extensive statistical analysis. The prompt set draws in part from established BBQ~\cite{Parrish2022} benchmark, which provides carefully controlled bias probes for question answering. Prompts span a broad range of demographic and contextual categories, including intersectional scenarios. Corpus composition and categorization were fixed prior to evaluation to ensure reproducibility and cross-model comparability.

\subsection{Dimensional decomposition}

SHARP decomposes social harm into four dimensions: bias ($B$), fairness ($F$), ethics ($E$), and epistemic reliability ($K$). For each dimension, a sub-index in $[0,1]$ is constructed by aggregating relevant judge outputs using root-mean-square (RMS) aggregation, which emphasizes large deviations and admits a geometric interpretation. Categorical attributes are mapped to ordered numeric scales. Full constructions are provided in Appendix~\ref{app:methodology}.

\subsection{Risk-sensitive judge aggregation}

To account for disagreement among judges, SHARP aggregates judge scores using a log-sum-exp (LSE) operator:
\begin{equation}
\mathrm{LSE}_\tau(X_1,X_2,X_3)
=
\tau \log\!\left(
\frac{1}{3}\sum_{j=1}^{3} e^{X_j/\tau}
\right),
\end{equation}
which smoothly interpolates between mean and max pooling. Smaller $\tau$ emphasizes worst-case assessments while retaining differentiability. We fix $\tau=0.20$, with sensitivity analysis reported in Appendix~\ref{appendix:lse_sensitivity}.

\subsection{Harm representation and compounded risk}
\label{sec:compounded-risk}

Each prompt-level response is represented as a four-dimensional harm vector
\[
h_{M,q} = (\bar{B}_{M,q}, \bar{F}_{M,q}, \bar{E}_{M,q}, \bar{K}_{M,q}) \in [0,1]^4,
\]
where each coordinate denotes the judge-aggregated sub-index for a distinct harm dimension.
The quantities $\bar{B}, \bar{F}, \bar{E}, \bar{K}$ are normalized \emph{harm intensities} derived from structured judge assessments; they are not calibrated probabilities of real-world harm.

SHARP aggregates multi-dimensional harm using a monotone \emph{compounding operator} inspired by union-of-failures models from reliability analysis. Specifically, we define a prompt-level \emph{compounded harm score}
\begin{equation}
H_{M,q} = 1 - \prod_{i \in \{B,F,E,K\}} (1 - h_{i,M,q}),
\label{eq:union}
\end{equation}
which increases whenever any dimension exhibits elevated harm and amplifies coordinated activation across dimensions. Equation~\eqref{eq:union} should be interpreted as an
\emph{operational risk aggregation}, not as a calibrated probability of harm occurrence.
The conditional independence implicit in the product form is a heuristic approximation that ensures monotonicity and interaction sensitivity; SHARP does not claim a probabilistic or causal interpretation of $H_{M,q}$.

To obtain an additive and decomposable risk representation, SHARP reparameterizes
residual safety via a negative log transform. For each dimension,
\[
\ell_{i,M,q} = -\log(1 - h_{i,M,q} + \varepsilon), \qquad \varepsilon = 10^{-6},
\]
and defines the \emph{cumulative log-risk}
\begin{equation}
L_{M,q} = \sum_{i} \ell_{i,M,q} = -\log(1 - H_{M,q} + \varepsilon).
\label{eq:logrisk}
\end{equation}
This transformation maps bounded harm intensities to unbounded risk units, preserves ordering, avoids saturation near one, and enables transparent attribution of compounded
risk to individual dimensions through the additive decomposition of $L_{M,q}$.

Throughout the paper, $H_{M,q}$ and $L_{M,q}$ are treated as \emph{ordinally meaningful risk scores}, rather than calibrated probabilities of real-world harm, whose primary role is comparative and distributional. SHARP’s core claims
concern relative tail behavior and structural differences across models under a fixed evaluation protocol, rather than calibrated estimation of real-world harm probabilities.

\subsection{Auxiliary magnitude diagnostic}

SHARP additionally reports a harm magnitude diagnostic defined as the Euclidean norm of the harm vector,
\begin{equation}
r_{M,q} = \|\mathbf{h}_{M,q}\|_2,
\end{equation}
which measures overall harm intensity without requiring joint activation and serves as a complementary indicator of prompt sensitivity and dispersion.

\subsection{Model-level risk profiling}

Each model induces an empirical distribution $\{L_{M,q}\}_{q\in\mathcal{Q}}$ of prompt-level cumulative log-risk values. SHARP summarizes this distribution using the expected log-risk $\mu_L(M)$, risk volatility $\sigma_L(M)$, and tail risk quantified via Conditional Value at Risk:
\begin{equation}
\mathrm{CVaR}_{0.95}(M)
=
\mathbb{E}\!\left[
L_{M,q} \mid
L_{M,q} \ge \mathrm{VaR}_{0.95}(M)
\right].
\end{equation}
CVaR$_{0.95}$ captures the severity of worst-case compounded failures and serves as SHARP’s primary safety-relevant statistic.

\paragraph{Methodological contribution.}
SHARP combines explicit dimensional decomposition, union-based harm aggregation, additive log-risk reparameterization, and distributional profiling via CVaR to expose failure modes and tail behaviors that are systematically obscured by mean-centered scalar benchmarks.

\section{Empirical Evaluation}
\label{sec:results}

\subsection{Model-Level Social Harm Risk Profiles}
\label{sec:harm-profiles}

We characterize model behavior using \textbf{CVaR$_{95}$ of cumulative log-risk} as the primary evaluation statistic, capturing worst-case \emph{compounded} social harm across bias, fairness, ethics, and epistemic dimensions. CVaR$_{95}$ measures the expected harm among the most adverse five percent of prompts and is therefore sensitive to tail behavior that mean-centered metrics systematically obscure. All reported statistics are computed over a shared set of $n = 901$ prompts.

Table~\ref{tab:key-findings-summary} summarizes model-level risk profiles using cumulative log-risk as the primary tail-sensitive metric, alongside complementary summaries of central tendency and probability-based tail behavior. Reported statistics include mean cumulative log-risk, CVaR${95}$ of cumulative log-risk to characterize worst-case compounded harm, and CVaR${95}$ of any-harm probability to illustrate saturation effects under probabilistic aggregation.

\begin{table*}[ht]
\centering
\caption{
Model-level summary of key SHARP metrics.
Lower values indicate lower estimated social harm.
CVaR$_{95}$ statistics characterize worst-case behavior across prompts.
}
\label{tab:key-findings-summary}
\setlength{\tabcolsep}{5pt}
\footnotesize
\begin{tabular}{lcccc}
\toprule
\textbf{Model} &
\textbf{CVaR$_{95}$ Cumulative Log-Risk} &
\textbf{Mean Cumulative Log-Risk} &
\textbf{Any-Harm CVaR$_{95}$} \\
\midrule
claude-sonnet-4.5  & 1.689 & 0.158 &  0.772 \\
gemini-1.5-pro     & 3.499 & 0.463 &  0.962 \\
claude-3.5-sonnet  & 3.782 & 0.449 &  0.965 \\
gemini-2.5-pro     & 3.868 & 0.560 &  0.971 \\
qwen3-235b         & 3.895 & 0.589 &  0.967 \\
gpt-4o             & 4.519 & 0.748 &  0.984 \\
gpt-oss-120b       & 5.286 & 1.209 &  0.994 \\
deepseek-chat      & 5.735 & 1.296 &  0.994 \\
mistral-large      & 5.988 & 1.322 &  0.995 \\
llama3-3-70b       & 7.934 & 1.968 &  0.998 \\
llama3-1-405b      & 8.397 & 2.154 &  0.998 \\
\bottomrule
\end{tabular}
\end{table*}

\textbf{Large divergence in worst-case risk.}
Models that appear similar under average risk exhibit pronounced separation in tail behavior. Claude Sonnet~4.5 attains a CVaR$_{95}$ of cumulative log-risk of $1.69$, whereas DeepSeek-Chat and Mistral-Large reach $5.73$ and $5.99$, corresponding to more than a threefold increase in worst-case compounded harm. At the upper end, LLaMA-3~405B exhibits a CVaR$_{95}$ of $8.40$, exceeding the lowest-risk model by over a factor of four. These separations are substantially larger than those observed under mean log-risk, which spans a comparatively narrow range.

\textbf{Mean risk understates tail exposure.}
Mean cumulative log-risk systematically underestimates extreme behavior across models. Gemini-1.5-Pro and Claude-3.5-Sonnet exhibit nearly identical mean log-risk values ($0.46$ versus $0.45$), yet differ meaningfully in CVaR$_{95}$ ($3.50$ versus $3.78$). This pattern recurs across the model set, demonstrating that mean-based summaries fail to capture worst-case failure modes that dominate deployment risk. Because low measured risk may reflect refusal or deflection rather than safe task completion, these results should be interpreted as intrinsic harm profiles conditional on observed outputs rather than utility-constrained performance.

\textbf{Probability metrics saturate under compounding.}
Probability-level metrics, such as any-harm probability, rapidly saturate in the tail. Most models exceed $0.96$ CVaR$_{95}$ in any-harm probability, with several approaching $0.99$. This saturation limits discriminative power and motivates cumulative log-risk as a more informative metric under compounding, where it remains well-spread and stable across models.

\textbf{Magnitude alone fails to capture interaction-driven risk.}
Geometric magnitude metrics, such as harm-radius CVaR$_{95}$, capture overall severity but fail to reflect amplification due to joint activation across dimensions. Models with similar harm-radius profiles can diverge substantially in cumulative log-risk, confirming that worst-case harm is driven by coordinated elevation across dimensions rather than marginal magnitude alone.

Collectively, these results demonstrate that worst-case social harm in LLMs is highly model-dependent, epistemically driven, and largely invisible to mean-based or probability-only evaluation. CVaR$_{95}$ of cumulative log-risk provides a stable and discriminative basis for comparing models under adverse prompt conditions.

\subsection{Interpreting SHARP for Risk-Constrained Model Selection}
\label{sec:decision-interpretation}

SHARP is intended for \emph{risk-constrained} model selection rather than total ordering by average performance.
We formalize this by framing evaluation as a decision problem under tail-risk tolerance.

Let $\mathcal{M}$ denote candidate models and let $L_M$ be the prompt-level cumulative log-risk distribution induced by model $M$.
For a specified tolerance $\tau > 0$, define the admissible set
\[
\mathcal{M}_\tau = \{ M \in \mathcal{M} : \mathrm{CVaR}_{0.95}(L_M) \le \tau \}.
\]
This criterion excludes models with unacceptable worst-case compounded harm regardless of mean risk.

As shown in Table~1, several models exhibit similar expected log-risk yet differ substantially in $\mathrm{CVaR}_{0.95}$.
Such models are indistinguishable under mean-centered evaluation but are separated under SHARP’s tail-sensitive criterion.
Conversely, models with slightly higher mean risk but lower tail exposure may be preferable in high-stakes settings. Rather than inducing a fragile total ranking, $\mathrm{CVaR}_{0.95}(L)$ supports filtering, exclusion, and tiered comparisons under explicit risk tolerance, aligning evaluation with deployment regimes in which rare but severe failures dominate expected harm.

\subsection{Risk Tiers and Statistical Indeterminacy}
\label{sec:risk-tiers}

Because SHARP operates on heavy-tailed prompt-level distributions, model comparison is best interpreted in terms of \emph{risk tiers} rather than strict total orderings.
Bootstrap confidence intervals for $\mathrm{CVaR}_{0.95}(L)$ (Appendix~G) show that while many model pairs are clearly separable, a subset of near-neighbors remains statistically indistinguishable.

Accordingly, SHARP induces a partial ordering: models with non-overlapping $\mathrm{CVaR}_{0.95}(L)$ intervals form distinct low- and high-risk tiers, while overlapping intervals define intermediate tiers where differences cannot be resolved under the current prompt distribution.
This structure is stable across bootstrap resampling, repeated-measures tests, and aggregation robustness analyses.

From a governance perspective, these tiers support conservative decision-making: higher-risk tiers can be excluded outright, while models within the same tier may be treated as interchangeable absent additional domain constraints.

\subsection{Decomposition of Harm into Four Spatial Dimensions}
\label{sec:subindex-decomposition}

Aggregate risk metrics obscure the structure of social harm.
SHARP therefore decomposes prompt-level harm into four dimensions: \emph{bias}, \emph{fairness}, \emph{ethical misalignment}, and \emph{epistemic unreliability}, to characterize heterogeneous failure modes that scalar summaries conflate.

Tables~\ref{tab:sharp_model_subindex_summary_bias_fairness}
and~\ref{tab:sharp_model_subindex_summary_ethics_epistemic}
report \emph{marginal} prompt-level means and tail severities (CVaR$_{95}$) for each sub-index.
These quantities describe how severe each harm dimension can become \emph{in isolation}, conditioning on dimension-specific tail events.
They are therefore descriptive of marginal failure behavior rather than determinants of dominance in compounded risk.

\begin{table*}[ht]
\centering
\caption{
Marginal decomposition of model-level SHARP harm profiles into bias and fairness sub-indices,
reporting prompt-level means and tail risk (CVaR$_{95}$).
Values correspond to expectations over prompts of log-sum-exp aggregated judge assessments
(Section~\ref{app:lse}); lower values indicate lower estimated harm.
}
\label{tab:sharp_model_subindex_summary_bias_fairness}
\setlength{\tabcolsep}{10pt}
\footnotesize
\begin{tabular}{@{}lcccc@{}}
\toprule
\textbf{Model} &
\textbf{Bias Mean} &
\textbf{Bias CVaR$_{95}$} &
\textbf{Fairness Mean} &
\textbf{Fairness CVaR$_{95}$} \\
\midrule
claude-sonnet-4.5  & 0.0210 & 0.3889 & 0.0352 & 0.2944 \\
gemini-1.5-pro     & 0.0734 & 0.6052 & 0.0869 & 0.5720 \\
claude-3.5-sonnet  & 0.0652 & 0.6317 & 0.0815 & 0.5855 \\
qwen3-235b         & 0.1037 & 0.6910 & 0.1124 & 0.6196 \\
gemini-2.5-pro     & 0.0997 & 0.7366 & 0.1050 & 0.6089 \\
gpt-4o             & 0.1256 & 0.7523 & 0.1368 & 0.6551 \\
gpt-oss-120b       & 0.1928 & 0.8179 & 0.2023 & 0.7326 \\
deepseek-chat      & 0.2183 & 0.8327 & 0.2216 & 0.7411 \\
mistral-large      & 0.2154 & 0.8401 & 0.2137 & 0.7493 \\
llama3-3-70b       & 0.3009 & 0.8828 & 0.3009 & 0.8043 \\
llama3-1-405b      & 0.3147 & 0.8881 & 0.3160 & 0.8102 \\
\bottomrule
\end{tabular}
\end{table*}

\begin{table*}[ht]
\centering
\caption{
Marginal decomposition of model-level SHARP harm profiles into ethics and epistemic
sub-indices, reporting prompt-level means and tail risk (CVaR$_{95}$).
Values correspond to expectations over prompts of log-sum-exp aggregated judge assessments
(Section~\ref{app:lse}); lower values indicate lower estimated harm.
}
\label{tab:sharp_model_subindex_summary_ethics_epistemic}
\setlength{\tabcolsep}{10pt}
\footnotesize
\begin{tabular}{@{}lcccc@{}}
\toprule
\textbf{Model} &
\textbf{Ethics Mean} &
\textbf{Ethics CVaR$_{95}$} &
\textbf{Epistemic Mean} &
\textbf{Epistemic CVaR$_{95}$} \\
\midrule
claude-sonnet-4.5  & 0.0244 & 0.2713 & 0.0523 & 0.3247 \\
gemini-1.5-pro     & 0.0687 & 0.4963 & 0.1177 & 0.6516 \\
claude-3.5-sonnet  & 0.0637 & 0.5116 & 0.1177 & 0.6959 \\
qwen3-235b         & 0.0846 & 0.5369 & 0.1280 & 0.6717 \\
gemini-2.5-pro     & 0.0837 & 0.5582 & 0.1165 & 0.6629 \\
gpt-4o             & 0.1129 & 0.6205 & 0.1606 & 0.7221 \\
gpt-oss-120b       & 0.1769 & 0.7323 & 0.2270 & 0.7774 \\
deepseek-chat      & 0.1848 & 0.7197 & 0.2424 & 0.7876 \\
mistral-large      & 0.1736 & 0.7177 & 0.2579 & 0.8069 \\
llama3-3-70b       & 0.2566 & 0.7797 & 0.3427 & 0.8511 \\
llama3-1-405b      & 0.2748 & 0.7962 & 0.4045 & 0.8857 \\
\bottomrule
\end{tabular}
\end{table*}

\textbf{Central tendency across harm dimensions.}
Beyond tail behavior, the sub-index means in Tables~\ref{tab:sharp_model_subindex_summary_bias_fairness} and \ref{tab:sharp_model_subindex_summary_ethics_epistemic}
reveal systematic differences in central tendency across the four harm dimensions.
Across all evaluated models, epistemic unreliability exhibits the highest mean values, followed by bias and fairness, with ethical misalignment consistently lowest.
Relative to bias, mean epistemic scores are elevated by approximately 10--60\% across models (roughly 25\% on average), indicating that epistemic issues arise more frequently across prompts.
Bias and fairness exhibit comparable mean magnitudes, while ethics occupies a uniformly lower central regime.
These mean-level patterns characterize the \emph{prevalence} of harm signals across typical prompts and should not be conflated with dominance in worst-case compounded risk, which we analyze separately via tail attribution.

\textbf{Interpretation of marginal tail severities.}
The tables above reveal substantial heterogeneity across dimensions.
Bias-related harm often exhibits higher marginal tail severity than epistemic unreliability for several models, while ethical misalignment occupies an intermediate regime.
Because each CVaR$_{95}$ conditions on a dimension-specific tail set, these quantities should be interpreted as characterizing isolated failure modes rather than identifying the dominant drivers of worst-case compounded harm.

\textbf{Attribution within compounded tail events.}
To determine which dimensions drive extreme \emph{compounded} failures, we perform a tail attribution analysis conditioned on the tail of cumulative log-risk $L$.
Specifically, for each model $M$ we define
$\mathcal{T}_M(0.95)=\{q: L_{M,q}\ge\mathrm{VaR}_{0.95}(L_M)\}$ and decompose $\mathrm{CVaR}_{0.95}(L_M)$ into additive log-risk contributions from each dimension.
Table~\ref{tab:sharp_tail_attribution} reports normalized contribution shares $S_i(M)$, which sum to one by construction.

This analysis shows that dominance in compounded tail risk is
\emph{model-dependent}. Epistemic unreliability constitutes a major contributor for several systems (e.g., Claude~3.5~Sonnet, GPT-4o), while bias-related mechanisms
dominate the compounded tail for others, including higher-risk
open-weight models. These findings clarify that dominance reflects persistent contribution across worst-case prompts rather than the largest marginal tail severity within any single dimension. Table~\ref{tab:sharp_tail_attribution} shows that the dominant contributors to compounded tail risk vary across models, underscoring that marginal sub-index tail severities alone are insufficient to characterize worst-case failure drivers.

\begin{table*}[ht]
\centering
\caption{
Tail attribution of compounded log-risk.
For each model $M$, we decompose the tail risk
$\mathrm{CVaR}_{0.95}(L_M)$ into dimension-wise additive
log-risk contributions.
Reported values are normalized shares
$S_i(M)=\mathbb{E}[\ell_i \mid q \in \mathcal{T}_M(0.95)] / \mathrm{CVaR}_{0.95}(L_M)$,
where $\mathcal{T}_M(0.95)$ denotes the top 5\% of prompts by cumulative
log-risk $L$.
By construction, $\sum_i S_i(M)=1$.
}
\label{tab:sharp_tail_attribution}
\setlength{\tabcolsep}{6pt}
\footnotesize
\begin{tabular}{@{}p{4.2cm}ccccc@{}}
\toprule
\textbf{Model} &
$\mathbf{\mathrm{CVaR}_{0.95}(L)}$ &
$\mathbf{S_B}$ &
$\mathbf{S_F}$ &
$\mathbf{S_E}$ &
$\mathbf{S_K}$ \\
\midrule
claude-sonnet-4-5        & 1.689 & 0.301 & 0.276 & 0.202 & 0.221 \\
gemini-1.5-pro               & 3.499 & 0.254 & 0.243 & 0.195 & 0.308 \\
claude-3-5-sonnet       & 3.782 & 0.255 & 0.229 & 0.190 & 0.327 \\
gemini-2.5-pro                   & 3.868 & 0.352 & 0.264 & 0.201 & 0.184 \\
qwen3-235b                       & 3.895 & 0.357 & 0.229 & 0.182 & 0.228 \\
gpt-4o                & 4.519 & 0.275 & 0.222 & 0.187 & 0.316 \\
gpt-oss-120b                     & 5.286 & 0.286 & 0.250 & 0.205 & 0.260 \\
deepseek-chat                    & 5.735 & 0.339 & 0.230 & 0.196 & 0.232 \\
mistral-large                    & 5.988 & 0.338 & 0.230 & 0.199 & 0.231 \\
llama3-3-70b                     & 7.934 & 0.435 & 0.200 & 0.166 & 0.188 \\
llama3-1-405b                    & 8.397 & 0.386 & 0.223 & 0.193 & 0.185 \\
\bottomrule
\end{tabular}
\end{table*}

\subsection{Statistical Validation}
\label{subsec:stat_validation}

We validate SHARP using complementary analyses that assess estimator stability, cross-model discriminability, and the relative contribution of model and prompt effects. Full statistical details are reported in the appendix.

\textbf{Estimator stability and uncertainty.}
We quantify uncertainty due to finite prompt sampling using paired non-parametric bootstrap resampling over prompts ($B=10{,}000$; Appendix~\ref{app:stat_validation}). For the primary outcome—prompt-level cumulative log-risk $L_{M,q}$ and its tail statistic $\mathrm{CVaR}_{0.95}(L)$—bootstrap confidence intervals are tight and well separated for most models. Pairwise bootstrap tests on $\Delta\mathrm{CVaR}_{0.95}(L)$ indicate that 80\% of model pairs are statistically separable at the 95\% level, with ambiguity concentrated among near-neighbor models. This pattern supports a tiered interpretation of risk rather than a fragile total ordering.

\textbf{Model-level heterogeneity.}
Because all models are evaluated on the same prompts, we apply the Friedman test, a non-parametric repeated-measures alternative to ANOVA, on paired prompt-level log-risk values. The omnibus test strongly rejects the null hypothesis of identical model behavior ($\chi^2(10)=1629.9$, $p\approx 0$), with a small-to-moderate effect size (Kendall’s $W=0.181$). Post-hoc Wilcoxon signed-rank tests with Holm correction find 87\% of model pairs remain significantly different, confirming that observed differences reflect systematic model effects rather than sampling noise, while again highlighting limited separability among adjacent models.

\textbf{Sources of variability.}
To contextualize effect sizes, we decompose variance in $L_{M,q}$ using a two-way fixed-effects model with model and prompt as factors (Appendix~\ref{app:variance_decomposition_new}). Prompt identity explains a larger share of total variance (25.8\%) than model identity (13.9\%), with the remainder attributable to residual and stochastic effects. This structure is expected in social-harm evaluation, which intentionally probes heterogeneous, context-sensitive scenarios. Model effects are statistically material but operate against a background of strong prompt dependence, motivating SHARP’s distributional and tail-risk focus.

\textbf{Robustness checks.}
Comparative conclusions are invariant to reasonable choices of hyperparameters. Model rankings induced by $\mathrm{CVaR}_{0.95}(L)$ are unchanged across LSE temperatures $\tau\in\{0.15,0.20,0.25\}$ and remain highly stable under variation of the CVaR tail threshold ($\alpha\in\{0.90,0.95,0.975\}$), indicating that results are not artifacts of aggregation or tail-selection choices (Appendix~\ref{appendix:lse_sensitivity}).

\textbf{Summary.}
Across bootstrap uncertainty analysis, non-parametric repeated-measures testing, variance decomposition, and robustness checks, SHARP yields stable, statistically distinguishable model risk profiles under the current compounded-risk definition. At the same time, persistent prompt-driven variability underscores why mean-centered metrics are insufficient and why distributional and tail-risk analysis is necessary for evaluating social harm in large language models.

\section{Limitations}
\label{sec:limitations}

SHARP is a diagnostic evaluation framework for characterizing distributional social harm under a fixed evaluation protocol, and several limitations constrain interpretation. First, SHARP’s harm sub-indices, compounded harm scores, and cumulative log-risk are \emph{operational risk measures}, not calibrated probabilities of real-world harm or normative ground-truth labels. Judge-derived harm intensities are ordinal and rubric-dependent; accordingly, all conclusions concern relative behavior, distributional structure, and tail exposure under a specified measurement configuration rather than absolute harm rates or deployment-ready safety guarantees. Cross-dimensional results should therefore be interpreted as comparative descriptions of marginal and compounded risk patterns induced by the evaluation protocol, rather than as cardinal claims about the intrinsic severity or real-world importance of specific harm dimensions.

Second, SHARP relies on LLM-based judges and inherits known limitations of LLM-as-a-judge paradigms, including correlated alignment effects, prompt sensitivity, and ambiguity in epistemic reliability assessment. Inter-judge disagreement is highest for epistemic harm, reflecting the intrinsic difficulty of evaluating factual soundness and uncertainty. Because the judge pool partially overlaps with the evaluated model set, reported scores and rankings are conditional on the chosen judge ensemble and may reflect systematic calibration biases. While robustness analyses demonstrate stability of relative tail-risk ordering under judge ablations, SHARP does not claim judge neutrality or independence, nor does it provide external human calibration in this study.

Third, measured social harm may be reduced by refusal, deflection, or minimal responses, which do not necessarily correspond to desirable behavior under real deployment constraints. SHARP evaluates intrinsic harm conditional on observed model outputs and does not explicitly model utility, coverage, or task completion; consequently, low estimated risk should not be interpreted as evidence of safe and useful performance absent additional capability or helpfulness constraints.

Fourth, SHARP evaluates single-turn intrinsic behavior under a fixed prompt corpus. The fairness dimension captures within-response inequitable treatment cues rather than population-level statistical fairness properties, which require aggregation across individuals or outcomes and are intentionally out of scope. Group-level fairness guarantees, downstream decision effects, and sociotechnical feedback loops are not modeled.

Finally, all reported risk profiles are conditional on the evaluated prompt distribution, which is primarily English-language and Western-centric. Prompt identity explains more variance than model identity, underscoring that SHARP induces conditional risk profiles rather than global model safety rankings. Tail-risk estimates such as CVaR$_{0.95}$ reflect the upper tail of the evaluated corpus under single-sample decoding; bootstrap uncertainty quantifies sampling variability, but distribution shift and generation stochasticity are not addressed in this study.

Overall, SHARP should be interpreted as a reproducible, distribution-aware framework for comparative tail-risk profiling under explicit measurement assumptions, not as a calibrated estimate of real-world harm or a substitute for context-specific validation in deployment settings.

\section{Conclusion}

This work shows that social harm in large language models is fundamentally \emph{distributional} and \emph{multidimensional}, and therefore poorly characterized by mean-centered scalar summaries. As LLMs are increasingly deployed in high-stakes settings, evaluation must account for rare but severe failures, prompt-dependent variability, and cross-dimensional interactions that conventional benchmarks obscure.

\textbf{Contributions.}
SHARP advances LLM evaluation as an operational risk-profiling framework along three axes. First, it decomposes social harm into bias, fairness, ethics, and epistemic reliability, enabling structured diagnosis of failure modes that scalar aggregates conflate. Second, it introduces a risk-sensitive aggregation via a union-of-failures formulation reparameterized as additive cumulative log-risk, yielding a decomposable measure aligned with worst-case behavior rather than average performance. Third, it characterizes model behavior using distributional statistics, with Conditional Value at Risk (CVaR$_{0.95}$) as a primary metric that explicitly surfaces tail exposure.

\textbf{Empirical findings.}
Across eleven frontier LLMs evaluated on a fixed corpus of socially sensitive prompts, SHARP demonstrates that models with similar mean risk can exhibit substantially different tail-risk profiles under the same evaluation protocol. In multiple cases, low expected risk coexists with elevated CVaR, revealing susceptibility to rare but severe failures that are invisible to mean-centered summaries and dominate worst-case exposure. Dimension-wise decomposition further shows heterogeneous marginal tail behavior across bias, fairness, ethics, and epistemic dimensions, with no single harm dimension uniformly driving tail outcomes across models. Statistical validation confirms that these differences reflect systematic model effects rather than sampling noise, while also highlighting strong prompt dependence in extreme-risk regimes.

\textbf{Implications for evaluation and governance.}
By treating harm as a multivariate random variable and prioritizing tail behavior, SHARP enables comparative assessments that are inaccessible to scalar benchmarks. Risk-sensitive thresholds support exclusion or tiering of models based on unacceptable worst-case exposure irrespective of mean performance, aligning evaluation with deployment regimes where tail events dominate practical risk. Dimensional decomposition further supports targeted mitigation by isolating dominant failure modes, without collapsing heterogeneous risks into a single scalar score.

\textbf{Limitations and outlook.}
SHARP evaluates intrinsic, single-turn behavior under a fixed prompt distribution and specified judge ensemble, and does not provide calibrated estimates of real-world harm or population-level guarantees. Reliance on LLM-based judges introduces rubric- and ensemble-conditionality, and low measured risk may reflect refusal or deflection absent explicit utility constraints. The evaluated corpus is primarily English-language and Western in context, and tail-risk estimates reflect single-sample decoding under the sampled prompt distribution. Future work should extend risk profiling to interactive and agentic settings, incorporate utility-aware constraints, broaden cultural and linguistic coverage, analyze generation stochasticity, and develop calibrated or hybrid judge-based assessment protocols.

In summary, SHARP provides a principled and reproducible foundation for comparative, tail-sensitive evaluation of social harm under explicit measurement assumptions. While not a calibrated measure of real-world impact, it demonstrates that meaningful assessment of LLM risk requires moving beyond scalar averages toward multidimensional, distribution-aware profiling of worst-case behavior.

\section*{Broader Impact}

This work introduces SHARP, a distribution-aware framework for evaluating social harm in large language models via multidimensional, tail-sensitive risk profiling. The primary goal of the framework is methodological: to improve how social harm is measured and compared under fixed evaluation protocols, particularly in regimes where rare but severe failures dominate practical risk.

A potential positive impact of this work is to enable more risk-aware model selection, auditing, and benchmarking in high-stakes deployment contexts. By exposing worst-case behavior and cross-dimensional failure structure that mean-centered metrics obscure, SHARP may support more conservative governance decisions, targeted mitigation strategies, and clearer communication of model risk profiles to practitioners and regulators.

At the same time, SHARP is not intended to estimate real-world harm rates, provide population-level fairness guarantees, or substitute for domain-specific validation. Misuse could arise if SHARP scores are interpreted as calibrated measures of societal impact, or if models are ranked or deployed solely on the basis of intrinsic harm profiles without considering task utility, deployment context, or affected stakeholders. To mitigate such risks, the framework explicitly emphasizes comparative, conditional interpretation under stated assumptions and highlights its limitations.

Overall, this work advances evaluation methodology for responsible machine learning by formalizing social harm as a multidimensional, distributional object. Its societal implications are primarily indirect, mediated through improved measurement and governance practices rather than direct deployment or decision-making.

\newpage
\bibliographystyle{plainnat}
\bibliography{bibliography}

\newpage

\appendix
\section{Methodology: Technical Details}
\label{app:methodology}

This appendix provides complete mathematical specifications for SHARP, expanding the summary in Section~\ref{sec:methodology}. Throughout, all normalized quantities lie in $[0,1]$, with larger values indicating greater estimated harm.

\subsection{Normalization of judge outputs}
\label{app:normalization}

Judge outputs are heterogeneous: boolean indicators, categorical labels, and ordinal scales each require distinct treatment. SHARP normalizes all judge outputs to the unit interval $[0,1]$, where larger values uniformly indicate greater social harm. This normalization enables commensurable combination across output types while preserving ordinality.

\paragraph{Ordinal scores.}
For any ordinal severity score $x \in \{1,2,\ldots,10\}$, the affine normalization
\begin{equation}
\phi_{\text{ord}}(x) = \frac{x - 1}{9}
\end{equation}
maps the minimum score to 0 and the maximum to 1. This transformation preserves linear spacing under the assumption of equal intervals between adjacent ratings.

\paragraph{Boolean indicators.}
For any boolean indicator $b \in \{\texttt{False}, \texttt{True}\}$,
\begin{equation}
\phi_{\text{bool}}(b) =
\begin{cases}
0 & \text{if } b = \texttt{False},\\
1 & \text{if } b = \texttt{True}.
\end{cases}
\end{equation}

\paragraph{Categorical variables.}
Certain categorical outputs admit natural orderings. For bias explicitness, which takes values in $\{\texttt{none}, \texttt{implicit}, \texttt{explicit}\}$, we employ the ordinal mapping
\begin{equation}
\phi_{\text{exp}}(t) =
\begin{cases}
0.0 & \text{if } t = \texttt{none},\\
0.5 & \text{if } t = \texttt{implicit},\\
1.0 & \text{if } t = \texttt{explicit}.
\end{cases}
\end{equation}
This trichotomous scale reflects that bias may manifest implicitly (subtle framing, omission, connotation) or explicitly (overt stereotyping or discriminatory statements). The mapping enforces monotonic escalation without introducing additional degrees of freedom.

\subsection{Sub-index construction}
\label{app:subindices}

SHARP decomposes social harm into four constituent dimensions, each capturing a distinct mechanism by which model outputs can cause societal harm: \emph{bias} ($B$), \emph{fairness} ($F$), \emph{ethics} ($E$), and \emph{epistemic reliability} ($K$).
For model $M$, prompt $q$, and judge $j$, we compute
\[
B_{M,q,j},\quad F_{M,q,j},\quad E_{M,q,j},\quad K_{M,q,j} \in [0,1].
\]
All sub-indices are constructed using fixed, interpretable coefficients chosen \emph{a priori} for transparency and controlled behavior rather than empirical optimization.

\subsubsection{Bias sub-index}

The bias sub-index models three facets of biased content: breadth of categories affected, intensity of harm, and amplifying factors.

\paragraph{Bias coverage (breadth).}
Let $\mathcal{C}=\{c_1,\ldots,c_k\}$ denote the set of $k$ bias category indicators (e.g., gender, race, ethnicity, disability, age, religion, geographic origin). Coverage quantifies the fraction of categories for which bias is detected:
\begin{equation}
\text{cov}_{M,q,j} =
\frac{1}{|\mathcal{C}|}\sum_{c\in\mathcal{C}} \phi_{\text{bool}}(c).
\end{equation}

\paragraph{Bias intensity.}
Let $s=\phi_{\text{ord}}(\texttt{bias\_severity})$ and $i=\phi_{\text{ord}}(\texttt{bias\_impact})$ denote normalized severity and impact scores. Intensity is computed as the RMS of these components:
\begin{equation}
\text{int}_{M,q,j} =
\sqrt{\frac{s^2+i^2}{2}}.
\end{equation}

\paragraph{Base bias.}
Coverage and intensity are combined with fixed weights:
\begin{equation}
B^{\text{base}}_{M,q,j} =
0.35\cdot \text{cov}_{M,q,j} + 0.65\cdot \text{int}_{M,q,j}.
\end{equation}
The heavier weight on intensity encodes the design choice that concentrated severe bias may be more consequential than diffuse mild bias.

\paragraph{Amplification factors.}
Two bounded modifiers, \texttt{explicitness} and \texttt{intersectionality}, escalate harm conditional on presence. Let $e=\phi_{\text{exp}}(\texttt{explicitness})$ and $t=\phi_{\text{bool}}(\texttt{intersectionality})$. Define
\begin{equation}
A_{M,q,j} = 1 + 0.25\cdot e + 0.25\cdot t,
\end{equation}
so $A_{M,q,j}\in[1,1.5]$.

\paragraph{Sub-index coefficients and design choices.}
All sub-index constructions in SHARP use fixed coefficients specified \emph{a priori}.
These coefficients are not learned, optimized, or claimed to be statistically optimal.
Instead, they encode transparent \emph{stakeholder risk preferences} about how different facets of harm should be emphasized when constructing an operational risk signal.

For example, in the bias sub-index, intensity is assigned greater weight than coverage because concentrated, severe bias affecting a small number of groups may plausibly be more consequential than diffuse but mild bias spread across categories. This weighting reflects a normative judgment about harm severity, not an empirical claim about real-world impact, and alternative stakeholders may reasonably adopt different weightings.

SHARP does not assert that any particular coefficient configuration is uniquely correct.
Rather, the framework separates \emph{measurement} from \emph{normative aggregation}:
the underlying judge-derived indicators remain fixed, while coefficients act as explicit governance parameters that can be adjusted to reflect institutional priorities, regulatory thresholds, or domain-specific risk tolerances.

\paragraph{Final bias sub-index.}
\begin{equation}
B_{M,q,j} = \min\!\bigl(1,\; B^{\text{base}}_{M,q,j}\cdot A_{M,q,j}\bigr).
\end{equation}

\subsection{Fairness sub-index}
\label{sec:fairness}

The fairness dimension in SHARP is intended to capture \emph{within-response inequitable
treatment cues} expressed in a single model output, rather than population-level statistical
fairness metrics that require aggregation across individuals or outcomes.
Accordingly, SHARP’s fairness sub-index is an intrinsic, prompt-level construct that
operationalizes fairness as a linguistic and normative property of the generated response.

For each prompt response pair, judges assess three normalized indicators:
(i) demographic disparity ($f_1$), reflecting differential treatment or standards applied to
explicitly referenced social groups within the response;
(ii) equality-of-opportunity gap ($f_2$), capturing denial, restriction, or conditional access to
resources, opportunities, or rights based on group membership; and
(iii) group disparity ($f_3$), reflecting hierarchical framing, normative privileging, or
systematic devaluation of particular groups.
Each indicator is scored based on the content of the response itself, without requiring
counterfactual generation or across-prompt aggregation.

The fairness sub-index is computed using root-mean-square aggregation,
\[
F_{M,q,j} = \sqrt{\frac{f_1^2 + f_2^2 + f_3^2}{3}},
\]
which emphasizes pronounced inequities while preserving boundedness and interpretability.

We emphasize that this formulation does not attempt to estimate demographic parity, equalized odds, or other statistical fairness criteria traditionally defined at the dataset or population level. Instead, SHARP’s fairness axis captures whether a \emph{single response} exhibits inequitable treatment signals that would plausibly contribute to social harm if encountered in deployment. Group-level fairness analysis remains an important but distinct problem and is intentionally out of scope for SHARP’s prompt-level evaluation setting.

\subsubsection{Ethics sub-index}

The ethics sub-index aggregates five normalized indicators of value misalignment: ethical misalignment ($g_1$), value conflict ($g_2$), harm risk ($g_3$), cultural insensitivity ($g_4$), and exclusion risk ($g_5$):
\begin{equation}
E_{M,q,j} = \sqrt{\frac{\sum_{i=1}^{5} g_i^2}{5}}.
\end{equation}

\subsubsection{Epistemic sub-index}

The epistemic sub-index aggregates two normalized indicators of unreliability: epistemic unsoundness ($k_1$) and epistemic risk ($k_2$):
\begin{equation}
K_{M,q,j} = \sqrt{\frac{k_1^2 + k_2^2}{2}}.
\end{equation}

\subsection{Geometric aggregation principle: Root-mean-square (RMS)}
\label{app:rms}

For a vector $\mathbf{x}=(x_1,\ldots,x_d)\in[0,1]^d$, the root-mean-square is
\begin{equation}
\mathrm{RMS}(\mathbf{x}) = \sqrt{\frac{1}{d}\sum_{i=1}^{d} x_i^2} = \frac{\|\mathbf{x}\|_2}{\sqrt{d}}.
\end{equation}
RMS preserves boundedness, emphasizes large deviations relative to arithmetic averaging, and admits a geometric interpretation as a normalized Euclidean distance in the unit hypercube. SHARP uses RMS both within sub-index components and as an auxiliary magnitude diagnostic over the four harm dimensions.

\subsection{Judge ensemble aggregation via log-sum-exp}
\label{app:lse}

Each prompt model pair is evaluated by multiple judges. Arithmetic averaging may attenuate rare but severe harm signals, while max pooling is brittle. SHARP employs log-sum-exp (LSE) aggregation as a risk-sensitive compromise. For a harm dimension $X\in\{B,F,E,K\}$ with judge-specific values $X_1,X_2,X_3\in[0,1]$ and temperature $\tau>0$:
\begin{equation}
\mathrm{LSE}_\tau(X_1,X_2,X_3)
=
\tau \log\!\left(\frac{1}{3}\sum_{j=1}^{3} e^{X_j/\tau}\right).
\end{equation}
LSE satisfies $\bar{x}\le \mathrm{LSE}_\tau(\mathbf{x})\le \max_j x_j$ and interpolates between mean pooling ($\tau\to\infty$) and max pooling ($\tau\to 0^+$). SHARP fixes $\tau=0.20$, with sensitivity analysis over $\tau\in\{0.15,0.20,0.25\}$ reported in Appendix~\ref{appendix:lse_sensitivity}. The resulting prompt-level judge-aggregated sub-indices are denoted $\bar{B}_{M,q}$, $\bar{F}_{M,q}$, $\bar{E}_{M,q}$, and $\bar{K}_{M,q}$.

\subsection{Harm-space embedding and compounded risk via additive log-risk}
\label{app:harm_embedding}

\paragraph{Embedding into harm space.}
For each model $M$ and prompt $q$, SHARP embeds the response into a four-dimensional harm space using the judge-aggregated sub-indices:
\begin{equation}
\mathbf{h}_{M,q} =
\bigl(\bar{B}_{M,q},\; \bar{F}_{M,q},\; \bar{E}_{M,q},\; \bar{K}_{M,q}\bigr)\in[0,1]^4.
\end{equation}
Higher coordinates indicate greater estimated harm along the corresponding dimension.

\paragraph{Auxiliary harm radius.}
As a magnitude diagnostic that does not require joint activation, SHARP reports the (normalized) Euclidean radius:
\begin{equation}
r_{M,q} = \mathrm{RMS}(\mathbf{h}_{M,q})
=
\sqrt{\frac{\bar{B}_{M,q}^2 + \bar{F}_{M,q}^2 + \bar{E}_{M,q}^2 + \bar{K}_{M,q}^2}{4}}
\in[0,1].
\end{equation}

\paragraph{Union-of-failures aggregated harm.}
To model the event that \emph{at least one} harm dimension is activated, SHARP defines the prompt-level aggregated harm probability
\begin{equation}
H^{\mathrm{any}}_{M,q}
=
1 - \prod_{i\in\{B,F,E,K\}} (1 - h_{i,M,q}),
\label{eq:any_harm_prob}
\end{equation}
where $h_{i,M,q}$ denotes the corresponding coordinate of $\mathbf{h}_{M,q}$.
Equation~\eqref{eq:any_harm_prob} is the standard union-of-failures construction. It assumes conditional independence across dimensions given $(M,q)$ as a first-order approximation for aggregation, not a causal claim.

\paragraph{Residual safety and additive log-risk.}
While $H^{\mathrm{any}}_{M,q}$ yields an interpretable probability of \emph{any} harm, its bounded scale can saturate near 1 and is less decomposable. SHARP therefore reparameterizes residual safety,
\[
S_{M,q} \;=\; 1 - H^{\mathrm{any}}_{M,q} \;=\; \prod_{i} (1 - h_{i,M,q}),
\]
using a negative log transform. Define dimension-wise log-risk contributions
\begin{equation}
\ell_{i,M,q} = -\log(1 - h_{i,M,q} + \varepsilon),
\qquad \varepsilon = 10^{-6},
\end{equation}
and the cumulative log-risk (equivalently, negative log residual safety)
\begin{equation}
L_{M,q} = \sum_{i\in\{B,F,E,K\}} \ell_{i,M,q}
=
-\log\!\bigl(1 - H^{\mathrm{any}}_{M,q} + \varepsilon\bigr).
\label{eq:cum_log_risk}
\end{equation}
This transformation preserves ordering (higher harm $\Rightarrow$ larger $L_{M,q}$), maps $h_i=0$ to $\ell_i=0$, and diverges as $h_i\to 1$, emphasizing near-maximal failures. Equation~\eqref{eq:cum_log_risk} also establishes that compounded risk is additive in log space, enabling transparent dimensional attribution via the summands $\ell_{i,M,q}$.

\paragraph{Interpretation.}
The quantity $-\log(1-h)$ corresponds to a cumulative hazard-style reparameterization of bounded harm into unbounded risk units. In this representation, independent failure pressures add, and tail events in any dimension contribute sharply to $L_{M,q}$, aligning the compounded metric with worst-case risk sensitivity.

\subsection{Model-level risk profiling and risk statistics}
\label{app:risk}

Each model $M$ induces an empirical distribution of prompt-level cumulative log-risk values $\{L_{M,q}\}_{q\in\mathcal{Q}}$ over the evaluation corpus. SHARP summarizes this distribution using complementary statistics capturing central tendency, dispersion, and tail behavior.

\paragraph{Expected log-risk.}
\begin{equation}
\mu_L(M) = \frac{1}{n}\sum_{q\in\mathcal{Q}} L_{M,q}.
\end{equation}

\paragraph{Log-risk volatility.}
\begin{equation}
\sigma_L(M) =
\sqrt{\frac{1}{n}\sum_{q\in\mathcal{Q}} \bigl(L_{M,q} - \mu_L(M)\bigr)^2 }.
\end{equation}
Higher volatility indicates prompt sensitivity and behavioral instability, corresponding to models that are typically lower-risk but exhibit severe compounded failures on a subset of prompts.

\paragraph{Tail risk via Conditional Value at Risk.}
For confidence level $\alpha=0.95$, define the Value at Risk (VaR) as the $\alpha$-quantile of $\{L_{M,q}\}$:
\begin{equation}
\mathrm{VaR}_\alpha(M) =
\inf\{x\in\mathbb{R}:\Pr(L_{M,q}\le x)\ge \alpha\},
\end{equation}
and Conditional Value at Risk (CVaR) as
\begin{equation}
\mathrm{CVaR}_\alpha(M) =
\mathbb{E}\!\left[L_{M,q}\mid L_{M,q}\ge \mathrm{VaR}_\alpha(M)\right].
\end{equation}
For $\alpha=0.95$, $\mathrm{CVaR}_{0.95}(M)$ measures the mean compounded log-risk among the worst-performing 5\% of prompts. CVaR is a coherent risk measure that captures tail severity rather than merely the onset of extreme behavior and serves as SHARP’s primary safety-relevant comparison metric.

\subsection{Matrix representation}
\label{app:matrix}

The per-judge harm vector is
\begin{equation}
\mathbf{z}_{M,q,j} =
(B_{M,q,j},\, F_{M,q,j},\, E_{M,q,j},\, K_{M,q,j})^\top \in [0,1]^4.
\end{equation}

The judge-aggregated harm matrix for model $M$ is
\begin{equation}
\mathbf{Z}_M =
\begin{bmatrix}
\mathbf{h}_{M,q_1}^\top\\
\vdots\\
\mathbf{h}_{M,q_n}^\top
\end{bmatrix}
\in [0,1]^{n\times 4},
\end{equation}
and the corresponding prompt-level cumulative log-risk vector is
\begin{equation}
\mathbf{L}_M = (L_{M,q_1}, \ldots, L_{M,q_n})^\top \in \mathbb{R}_+^{n}.
\end{equation}

Dimension-specific expected harms are given by
\begin{equation}
\mathbf{H}_M =
\frac{1}{n}\mathbf{1}^\top \mathbf{Z}_M
=
(\mathbb{E}[\bar{B}],\, \mathbb{E}[\bar{F}],\, \mathbb{E}[\bar{E}],\, \mathbb{E}[\bar{K}])^\top.
\end{equation}
For a policy weight vector $\mathbf{w}\in\Delta^3$ (the 3-simplex), a linear policy-weighted summary is
\begin{equation}
s_M = \mathbf{w}^\top \mathbf{H}_M.
\end{equation}
This separation of measurement ($\mathbf{Z}_M$) from normative weighting ($\mathbf{w}$) supports transparent governance and stakeholder-specific aggregation.

\subsection{Methodological contributions (technical view)}
\label{app:contrib}

SHARP contributes the following methodological components relative to mean-centered scalar evaluation:
\begin{enumerate}
\item \textbf{Dimensional decomposition.} Harm is decomposed into bias, fairness, ethics, and epistemic dimensions, enabling targeted diagnosis and transparent reporting.
\item \textbf{Risk-sensitive aggregation.} Judge ensembling via LSE emphasizes elevated harm assessments while remaining robust to single-judge outliers.
\item \textbf{Compounded risk via additive log-risk.} Prompt-level multi-dimensional harm is aggregated through a union-of-failures construction and reparameterized as cumulative log-risk, yielding an additive, decomposable compounded risk signal aligned with worst-case behavior.
\item \textbf{Distributional profiling.} Reporting mean, volatility, and CVaR$_{0.95}$ over prompt-level risk distributions characterizes both typical behavior and tail failures.
\end{enumerate}

\section{Judge Agreement and Disagreement Structure}
\label{app:judge_agreement}

This appendix characterizes the reliability and disagreement structure of the three-judge protocol used throughout SHARP. We report (i) dispersion of judge assessments via mean absolute deviation (MAD) computed at the prompt level within each harm dimension, (ii) stability of model tail-risk rankings under leave-one-judge-out (LOJO) re-aggregation using CVaR$_{0.95}$, and (iii) prompt-level concordance between judges via pairwise Kendall's $\tau$ computed separately for each model and harm dimension. These diagnostics are intended to quantify sensitivity to judge identity and to separate systematic disagreements from idiosyncratic noise.

\paragraph{Evaluation setting and configuration.}
All results in this appendix use $n=901$ prompts evaluated across $|\mathcal{M}|=11$ models with $K=3$ blinded LLM judges. Judge scores are aggregated with log-sum-exp using temperature $\tau=0.2$. For tail-risk summaries, we report CVaR$_{0.95}$. Pairwise Kendall's $\tau$ is computed only when the overlap between two judges' prompt-level scores is at least 25 prompts per model and dimension. (Complete configuration and provenance are recorded in the accompanying generated report.)%

\subsection{Inter-judge dispersion via mean absolute deviation}
\label{app:judge_mad}

Let $s_{q,M}^{(k,d)} \in [0,1]$ denote judge $k$'s score for prompt $q$ and model $M$ under dimension $d \in \{\text{bias, fairness, ethics, epistemic}\}$. For each $(q,M,d)$ we compute the mean absolute deviation across judges,
\[
\mathrm{MAD}_{q,M}^{(d)} \;=\; \frac{1}{K}\sum_{k=1}^{K}\left|s_{q,M}^{(k,d)} - \frac{1}{K}\sum_{k'=1}^{K}s_{q,M}^{(k',d)}\right|.
\]

Table~\ref{tab:appE_mad_by_dimension} reports the mean and standard deviation of $\mathrm{MAD}_{q,M}^{(d)}$ aggregated over all evaluated prompt model pairs (total $n=9{,}911$ per dimension). Lower MAD indicates tighter inter-judge agreement. Inter-judge dispersion varies systematically across harm dimensions, with epistemic assessments exhibiting the largest average MAD, followed by fairness and bias, and ethics exhibiting the lowest dispersion. These differences reflect dimension-dependent variability in judge scoring behavior under the current rubric, rather than calibrated differences in underlying harm severity.

\begin{table}[t]
\centering
\caption{Inter-judge dispersion by dimension via mean absolute deviation (MAD). MAD is computed per (prompt, model) across the three judges and summarized over all evaluated prompt model pairs ($n=9{,}911$ per dimension).}
\label{tab:appE_mad_by_dimension}
\setlength{\tabcolsep}{6pt}
\begin{tabular}{@{}lrrr@{}}
\toprule
\textbf{Dimension} & \textbf{$n$ items} & \textbf{MAD mean} & \textbf{MAD std} \\
\midrule
Ethics    & 9{,}911 & 0.0378 & 0.0585 \\
Bias      & 9{,}911 & 0.0447 & 0.0796 \\
Fairness  & 9{,}911 & 0.0460 & 0.0698 \\
Epistemic & 9{,}911 & 0.0568 & 0.0679 \\
\bottomrule
\end{tabular}
\end{table}

\subsection{Rank stability under leave-one-judge-out}
\label{app:judge_lojo}

We next test whether model ordering under tail risk is sensitive to the composition of the judge panel. For each LOJO condition, we remove one judge and re-aggregate the remaining two judges' scores using the same log-sum-exp operator at $\tau=0.2$, then compute model-level CVaR$_{0.95}$ over prompts using the same scalar harm construction as in the main pipeline for this analysis run. We compare the induced model ranking to the full three-judge ranking using Kendall's $\tau$ on ranks (Table~\ref{tab:appE_lojo_rank_stability}). The rankings are highly stable: removing the OpenAI judge yields identical ordering ($\tau=1.0$), while removing either of the other judges yields $\tau \approx 0.891$. This indicates that the CVaR-based model ordering is largely robust to single-judge ablations under the specified aggregation temperature and prompt distribution.

\begin{table}[t]
\centering
\caption{Leave-one-judge-out (LOJO) stability of model CVaR$_{0.95}$ rankings. We compute Kendall's $\tau$ between the full three-judge CVaR ranking and the LOJO ranking after omitting the specified judge.}
\label{tab:appE_lojo_rank_stability}
\setlength{\tabcolsep}{6pt}
\begin{tabular}{@{}lrrr@{}}
\toprule
\textbf{Judge omitted} & \textbf{$n$ models} & \textbf{Kendall $\tau$} & \textbf{$p$-value} \\
\midrule
OpenAI::gpt-5.1            & 11 & 1.0000 & $5.01\times10^{-8}$ \\
Anthropic::claude-sonnet-4-5 & 11 & 0.8909 & $1.37\times10^{-5}$ \\
Google::gemini-2.5-pro                & 11 & 0.8909 & $1.37\times10^{-5}$ \\
\bottomrule
\end{tabular}
\end{table}

\subsection{Prompt-level judge concordance}
\label{app:judge_kendall}

We next measure prompt-level concordance between judges within each harm dimension. For each model $M$ and dimension $d$, we compute Kendall's $\tau$ between each judge pair using their prompt-level score vectors $\{s_{q,M}^{(k,d)}\}_{q\in\mathcal{Q}}$, restricted to prompts with non-missing overlap. We then summarize these per-model $\tau$ values across the 11 evaluated models by their mean, median, and standard deviation (Table~\ref{tab:appE_pairwise_kendall}). Larger $\tau$ indicates stronger agreement in the relative ordering of prompts by estimated harm severity.

Across dimensions, we observe moderate prompt-level concordance between judges, with systematic variation across judge pairs and harm categories. In several dimensions, the Claude Gemini pair exhibits higher average concordance than pairs involving GPT-based judges, while other pairings show lower but still positive agreement. These patterns indicate dimension- and pair-specific differences in prompt-level ranking behavior rather than uniformly noisy or unstable assessments.

\begin{table}[t]
\centering
\caption{Pairwise prompt-level judge concordance via Kendall's $\tau$, aggregated over models. For each model and dimension, we compute Kendall's $\tau$ between a judge pair over prompt-level sub-index scores, then summarize $\tau$ across the 11 evaluated models.}
\label{tab:appE_pairwise_kendall}
\setlength{\tabcolsep}{4.5pt}
\begin{tabular}{@{}llrrrr@{}}
\toprule
\textbf{Dimension} & \textbf{Judge pair} & \textbf{$n$ models} & \textbf{Mean $\tau$} & \textbf{Median $\tau$} & \textbf{Std $\tau$} \\
\midrule
Bias & claude-sonnet-4-5 vs gemini-2.5-pro     & 11 & 0.6980 & 0.7127 & 0.0761 \\
Bias & claude-sonnet-4-5 vs gpt-5.1 & 11 & 0.5626 & 0.5774 & 0.0541 \\
Bias & gemini-2.5-pro vs gpt-5.1                & 11 & 0.5462 & 0.5518 & 0.0570 \\
\midrule
Fairness & claude-sonnet-4-5 vs gemini-2.5-pro     & 11 & 0.6536 & 0.6830 & 0.0924 \\
Fairness & claude-sonnet-4-5 vs gpt-5.1 & 11 & 0.5978 & 0.5928 & 0.0542 \\
Fairness & gemini-2.5-pro vs gpt-5.1                & 11 & 0.5344 & 0.5316 & 0.0940 \\
\midrule
Ethics & claude-sonnet-4-5 vs gemini-2.5-pro     & 11 & 0.6333 & 0.6522 & 0.0878 \\
Ethics & claude-sonnet-4-5 vs gpt-5.1 & 11 & 0.5557 & 0.5637 & 0.0501 \\
Ethics & gemini-2.5-pro vs gpt-5.1                & 11 & 0.5357 & 0.5394 & 0.0808 \\
\midrule
Epistemic & claude-sonnet-4-5 vs gpt-5.1 & 11 & 0.6428 & 0.6440 & 0.0370 \\
Epistemic & claude-sonnet-4-5 vs gemini-2.5-pro     & 11 & 0.5595 & 0.5566 & 0.1014 \\
Epistemic & gemini-2.5-pro vs gpt-5.1                & 11 & 0.4619 & 0.4452 & 0.1082 \\
\bottomrule
\end{tabular}
\end{table}

\paragraph{Implications for SHARP validity claims.}
Taken together, these diagnostics indicate that inter-judge agreement is dimension-dependent and that judges exhibit consistent but imperfect concordance in ranking prompts by harm severity. Combined with the leave-one-judge-out analyses, these results show that SHARP’s model-level tail-risk rankings remain stable with respect to judge identity under the chosen aggregation temperature, even in the presence of prompt-level ranking disagreements. Accordingly, SHARP’s comparative conclusions should be interpreted as robust to reasonable variation in judge composition, while recognizing that prompt-level harm ordering is inherently subject to rubric- and judge-specific variation.

\section{Judge Ensemble Sensitivity and Overlap Considerations}
\label{app:judge_sensitivity}

Judge model overlap is a known concern in LLM-as-a-judge evaluation, particularly when judges and evaluated models share training distributions or architectural lineage. SHARP is explicitly designed to accommodate alternative judge configurations without modification to its core metric definitions.

Several robustness analyses are natural extensions of the present study. First, a leave-one-judge-out protocol can be applied, recomputing all SHARP statistics using each two-judge subset to assess sensitivity to individual judges. Second, a fully disjoint judge ensemble can be substituted to eliminate self-evaluation effects entirely, enabling direct quantification of overlap-induced shifts in absolute harm levels and relative rankings. Third, cross-ensemble comparisons can be used to characterize variance attributable to judge choice, complementing prompt-level variance decomposition.

These analyses are orthogonal to SHARP’s formulation and can be executed without altering the harm taxonomy, geometric embedding, or risk-sensitive aggregation. We leave a comprehensive cross-judge sensitivity study to future benchmark instantiations as larger and more diverse judge pools become available.

\section{Empirical Test of the Independence Assumption}
\label{app:independence_test}

A central modeling assumption underlying harm representation  is that the four harm dimensions: bias ($B$), fairness ($F$), ethics ($E$), and epistemic reliability ($K$), are not perfectly dependent at the prompt level. While SHARP does not require strict independence, excessive correlation among sub-indices would undermine the interpretability of multiplicative aggregation and weaken the motivation for modeling joint harm via a union-of-failures formulation. To make this assumption empirically testable, we directly measure prompt-level dependence among $(B,F,E,K)$ using the materialized harm embeddings.

\paragraph{Methodology.}
For each evaluated model, we compute pairwise correlations among the four sub-indices across prompts using the prompt-level embeddings stored in \texttt{sharp\_harm\_space\_embedding\_v6}. Specifically, for each model $M$, we compute correlations for all six unordered pairs in $\{(B,F),(B,E),(B,K),(F,E),(F,K),(E,K)\}$. Correlations are computed using Spearman’s $\rho$, which captures monotonic dependence and is robust to non-Gaussian score distributions. We report, for each model, the mean correlation across the six pairs. 

To probe dependence in the regime most relevant for risk aggregation, we additionally compute \emph{tail-slice correlations} restricted to the top 5\% of prompts by cumulative log risk $L$. This tail slice isolates high-risk prompts where compounding effects would be most consequential and where violations of approximate independence would most strongly affect SHARP’s tail-risk statistics.

\paragraph{Results.}

\begin{table}[t]
\centering
\caption{Empirical dependence among harm dimensions. For each model, we compute prompt-level correlations among $(B,F,E,K)$ across all prompts and within the tail slice (top 5\% by $L$). Reported values are the mean correlation across the six dimension pairs.}
\label{tab:independence_test_correlations}
\setlength{\tabcolsep}{6pt}
\begin{tabular}{@{}p{5.2cm}rr@{}}
\toprule
\textbf{Evaluated model} & \textbf{Mean $\rho$ (all)} & \textbf{Mean $\rho$ (top 5\% by $L$)} \\
\midrule
claude-sonnet-4-5 & 0.638 & 0.483 \\
claude-3-5-sonnet & 0.743 & 0.448 \\
gemini-1.5-pro & 0.765 & 0.297 \\
qwen3-235b & 0.814 & 0.372 \\
gemini-2.5-pro & 0.814 & 0.246 \\
gpt-4o & 0.819 & 0.283 \\
llama3-1-405b & 0.857 & 0.113 \\
gpt-oss-120b & 0.859 & 0.012 \\
llama3-3-70b & 0.884 & 0.067 \\
deepseek-chat & 0.887 & 0.224 \\
mistral-large & 0.894 & 0.135 \\
\bottomrule
\end{tabular}
\end{table}

Table~\ref{tab:independence_test_correlations} summarizes the results. Across all prompts, mean pairwise correlations are moderate to high for most models, indicating that harm dimensions are not statistically independent in general. This is expected, as many socially sensitive prompts activate multiple forms of harm simultaneously. However, when restricting attention to the high-risk tail (top 5\% by $L$), correlations consistently decrease—often substantially—across all models. Several models exhibit near-zero mean correlations in the tail slice, suggesting that extreme-risk prompts are characterized by more heterogeneous and asymmetric activation patterns across dimensions rather than uniformly elevated harm.

\paragraph{Implications for Eq.~(3).}
These findings support the use of a multiplicative, union-of-failures–style aggregation despite moderate global dependence among sub-indices. While $(B,F,E,K)$ are correlated on average, the reduced dependence observed in the high-risk tail implies that worst-case harm is frequently driven by dimension-specific failures rather than fully coupled effects. Consequently, Eq.~(3) should be interpreted as an operational risk aggregation under \emph{approximate} independence in the regime that dominates tail risk, rather than as a claim of strict statistical independence across all prompts. This empirical validation strengthens the methodological grounding of SHARP without requiring explicit copula modeling or stronger parametric assumptions.

\section{LSE Temperature Sensitivity Analysis}
\label{appendix:lse_sensitivity}

SHARP aggregates three judge scores for each $(q,M)$ evaluation using log-sum-exp (LSE). The temperature $\tau$ controls the aggregation regime:
\begin{equation}
\mathrm{LSE}(\mathbf{s}; \tau)
= \tau \log\!\left(\frac{1}{K}\sum_{k=1}^{K} \exp\!\left(\frac{s_k}{\tau}\right)\right),
\qquad K=3,
\end{equation}
where $\mathbf{s}=(s_1,\ldots,s_K)$ are judge scores for a fixed harm dimension. Smaller $\tau$ increases sensitivity to the most severe judge assessment, while larger $\tau$ approaches arithmetic averaging and attenuates single-judge peaks.

\paragraph{Sensitivity design and metric.}
We evaluate robustness over $\tau \in \{0.15, 0.20, 0.25\}$ using SHARP's current compounded-risk construction: \emph{cumulative log-risk} $L_{M,q}$ computed after judge ensembling, and model-level ordering induced by $\mathrm{CVaR}_{0.95}(L)$ (lower is safer). We use $\tau=0.20$ as the reference setting.

\paragraph{Rank stability.}
Model rankings are \emph{invariant} across the tested temperature grid. Rank correlations with the reference ordering at $\tau=0.20$ are perfect (Kendall's $\tau_b=1.0$ and Spearman's $\rho=1.0$ for $\tau=0.15$ and $\tau=0.25$). This indicates that comparative conclusions are not an artifact of a particular temperature choice.

\begin{table}[ht]
\centering
\caption{Rank stability of model ordering induced by $\mathrm{CVaR}_{0.95}(L)$ across LSE temperatures, measured against the reference ordering at $\tau=0.20$.}
\label{tab:lse_rank_stability}
\small
\begin{tabular}{lccc}
\toprule
\textbf{$\tau$} & \textbf{$\tau_{\mathrm{ref}}$} & \textbf{Kendall $\tau_b$} & \textbf{Spearman $\rho$} \\
\midrule
0.15 & 0.20 & 1.000 & 1.000 \\
0.20 & 0.20 & 1.000 & 1.000 \\
0.25 & 0.20 & 1.000 & 1.000 \\
\bottomrule
\end{tabular}
\end{table}

\paragraph{Extremes and qualitative stability.}
The identity and ordering of the safest and riskiest models remain unchanged across $\tau$. Table~\ref{tab:lse_extremes_by_tau} reports the top-5 safest and bottom-5 riskiest models under $\mathrm{CVaR}_{0.95}(L)$ at each temperature, alongside mean log-risk. Both the head and tail of the ordering are stable, which is the safety-relevant regime for SHARP's risk profiling.

\begin{table}[ht]
\centering
\caption{Top-5 safest and bottom-5 riskiest models by $\mathrm{CVaR}_{0.95}(L)$ across LSE temperatures. Lower is safer.}
\label{tab:lse_extremes_by_tau}
\small
\setlength{\tabcolsep}{4pt}
\begin{tabular}{llcc}
\toprule
\textbf{$\tau$} & \textbf{Model} & \textbf{Mean log-risk} & \textbf{CVaR$_{0.95}$(log-risk)} \\
\midrule
\multicolumn{4}{l}{\textbf{Safest (lowest CVaR)}}\\
0.15 & claude-sonnet-4-5    & 0.156128 & 1.68009 \\
0.15 & gemini-1.5-pro            & 0.458956 & 3.46838 \\
0.15 & claude-3-5-sonnet    & 0.445173 & 3.75266 \\
0.15 & gemini-2.5-pro                 & 0.556568 & 3.85193 \\
0.15 & qwen3-235b & 0.584740 & 3.87204 \\
\midrule
0.20 & claude-sonnet-4-5    & 0.157465 & 1.68902 \\
0.20 & gemini-1.5-pro            & 0.462903 & 3.49870 \\
0.20 & claude-3-5-sonnet    & 0.448861 & 3.78177 \\
0.20 & gemini-2.5-pro                 & 0.560106 & 3.86761 \\
0.20 & qwen3-235b & 0.588877 & 3.89523 \\
\midrule
0.25 & claude-sonnet-4-5    & 0.159029 & 1.70089 \\
0.25 & gemini-1.5-pro            & 0.467818 & 3.53697 \\
0.25 & claude-3-5-sonnet    & 0.453398 & 3.81934 \\
0.25 & gemini-2.5-pro                 & 0.564731 & 3.89111 \\
0.25 & qwen3-235b & 0.594279 & 3.92749 \\
\midrule
\multicolumn{4}{l}{\textbf{Riskiest (highest CVaR)}}\\
0.15 & gpt-oss-120b          & 1.20068 & 5.25536 \\
0.15 & deepseek-chat                    & 1.28478 & 5.70648 \\
0.15 & mistral-large             & 1.31141 & 5.95198 \\
0.15 & llama3-3-70b  & 1.95353 & 7.90593 \\
0.15 & llama3-1-405b & 2.13724 & 8.35638 \\
\midrule
0.20 & gpt-oss-120b          & 1.20908 & 5.28608 \\
0.20 & deepseek-chat                    & 1.29552 & 5.73468 \\
0.20 & mistral-large             & 1.32181 & 5.98760 \\
0.20 & llama3-3-70b  & 1.96802 & 7.93393 \\
0.20 & llama3-1-405b & 2.15399 & 8.39657 \\
\midrule
0.25 & gpt-oss-120b          & 1.22001 & 5.32640 \\
0.25 & deepseek-chat                    & 1.30957 & 5.77213 \\
0.25 & mistral-large             & 1.33531 & 6.03271 \\
0.25 & llama3-3-70b  & 1.98689 & 7.96984 \\
0.25 & llama3-1-405b & 2.17555 & 8.44789 \\
\bottomrule
\end{tabular}
\end{table}

\paragraph{Cardinal sensitivity.}
Although ordinal conclusions are perfectly stable, absolute CVaR values change modestly with $\tau$, reflecting the intended behavior of LSE as it interpolates between more pessimistic and more averaging aggregation. Across models, the maximum spread in $\mathrm{CVaR}_{0.95}(L)$ over $\tau \in \{0.15,0.20,0.25\}$ is small relative to cross-model separations; the largest observed $\Delta$CVaR is 0.0915 (llama3-1-405b), and the smallest is 0.0208 (claude-sonnet-4-5-20250929). Table~\ref{tab:lse_cvar_spread} reports the per-model maximum CVaR spread across $\tau$.

\begin{table}[ht]
\centering
\caption{Per-model sensitivity magnitude: maximum $\Delta$CVaR$_{0.95}$(log-risk) across $\tau \in \{0.15, 0.20, 0.25\}$.}
\label{tab:lse_cvar_spread}
\small
\begin{tabular}{lc}
\toprule
\textbf{Model} & \textbf{Max $\Delta$CVaR across $\tau$} \\
\midrule
llama3-1-405b & 0.0915134 \\
mistral-large             & 0.0807286 \\
gpt-oss-120b          & 0.0710353 \\
gemini-1.5-pro              & 0.0685905 \\
claude-3-5-sonnet      & 0.0666781 \\
gpt-4o                & 0.0658673 \\
deepseek-chat                    & 0.0656494 \\
llama3-3-70b  & 0.0639092 \\
qwen3-235b   & 0.0554545 \\
gemini-2.5-pro                   & 0.0391798 \\
claude-sonnet-4-5      & 0.0207980 \\
\bottomrule
\end{tabular}
\end{table}

\paragraph{Interpretation.}
This analysis separates \emph{ordinal robustness} from \emph{cardinal sensitivity}. Under SHARP's current compounded-risk definition (cumulative log-risk after ensembling), model ordering by tail risk is invariant across a reasonable temperature range, while absolute CVaR magnitudes exhibit mild, structured variation. Since SHARP is designed for comparative risk assessment and tail profiling, the invariance of $\mathrm{CVaR}_{0.95}(L)$-based rankings provides strong evidence that the choice $\tau=0.20$ does not induce cherry-picked outcomes.

\section{Aggregation Robustness and Comparison to Simpler Risk Aggregates}
\label{app:aggregation_robustness}

This appendix evaluates the robustness of SHARP’s tail-risk aggregation by comparing it to several simpler, commonly used risk aggregates, without re-running model inference. All results are computed post hoc from the materialized prompt-level metrics in \texttt{sharp\_harm\_space\_embedding\_v6}, ensuring that differences arise solely from aggregation choice rather than evaluation noise or sampling variation.

\paragraph{Baseline aggregates.}
For each evaluated model, we compute four alternative model-level risk summaries over the same prompt distribution: (i) CVaR$_{0.95}$ of cumulative log-risk $L$ (SHARP’s primary statistic), (ii) CVaR$_{0.95}$ of the harm radius, (iii) CVaR$_{0.95}$ of the maximum sub-index $\max(B,F,E,K)$, and (iv) the mean any-harm probability. Table~\ref{tab:aggregation_baseline_metrics} (derived from the accompanying artifact) reports these quantities alongside the mean of $L$. As expected, mean-based summaries are substantially less sensitive to extreme failures, while CVaR-based aggregates amplify tail behavior.

\paragraph{Rank consistency across aggregates.}
To assess whether SHARP’s conclusions depend on the specific choice of aggregate, we compute rank correlations between model orderings induced by each summary statistic. As shown in Table~\ref{tab:aggregation_rank_correlations}, CVaR$_{0.95}(L)$ is highly correlated with other tail-sensitive aggregates, including CVaR of harm radius and CVaR of the maximum sub-index (Spearman $\rho \geq 0.95$ in all cases). Correlation with the mean any-harm probability is also strong, indicating that SHARP’s tail-risk ranking is broadly consistent with simpler baselines. At the same time, these correlations are not perfect, leaving room for meaningful reordering when tail behavior diverges.

\paragraph{Concrete decision flip.}
Despite high overall rank agreement, simpler aggregates can obscure practically relevant differences. The final block of results identifies a concrete decision flip between two models with nearly identical mean cumulative log-risk but materially different CVaR$_{0.95}(L)$. In this pair, the absolute difference in mean $L$ is negligible, yet the CVaR differs by approximately $0.45$, leading to a different risk ordering under tail-sensitive evaluation. This example demonstrates that mean-only aggregation can mask rare but severe failures that SHARP is explicitly designed to surface.

\paragraph{Implications.}
Taken together, these results show that SHARP is not an idiosyncratic aggregation that contradicts simpler baselines. Instead, it refines them by focusing the evaluation on the regime that dominates societal risk.High rank correlation establishes robustness, while the observed decision flip illustrates why tail-aware aggregation provides additional discriminative power beyond mean-based or max-based summaries. This comparison strengthens the empirical case for using CVaR of cumulative log-risk as SHARP’s primary model-level statistic. 

\subsection{Comparison to Simpler Risk Aggregates}
\label{app:comparision_aggregation_robustness}

All results in this subsection, shown in ~\ref{tab:aggregation_baseline_metrics}, ~\ref{tab:aggregation_rank_correlations}, and ~\ref{tab:aggregation_decision_flip} are computed post hoc, without re-running inference.

\begin{table}[t]
\centering
\caption{
Model-level risk aggregates computed over the same prompt distribution ($n=901$ per model).
Lower values indicate lower estimated risk. CVaR$_{0.95}$ denotes tail risk at the 95th percentile.
}
\label{tab:aggregation_baseline_metrics}
\setlength{\tabcolsep}{4pt}
\footnotesize
\begin{tabular}{@{}p{3cm}rrrrrr@{}}
\toprule
\textbf{Evaluated model} &
\textbf{Mean $L$} &
\textbf{CVaR$_{0.95}(L)$} &
\textbf{CVaR$_{0.95}(R)$} &
\textbf{CVaR$_{0.95}(\max)$} &
\textbf{Mean any-harm} \\
\midrule
claude-sonnet-4-5        & 0.157 & 1.689 & 0.344 & 0.438 & 0.101 \\
claude-3-5-sonnet        & 0.449 & 3.782 & 0.590 & 0.749 & 0.202 \\
gemini-1.5-pro                & 0.463 & 3.499 & 0.569 & 0.701 & 0.212 \\
gemini-2.5-pro                    & 0.560 & 3.868 & 0.600 & 0.752 & 0.232 \\
qwen3-235b    & 0.589 & 3.895 & 0.586 & 0.718 & 0.252 \\
gpt-4o                 & 0.747 & 4.519 & 0.653 & 0.812 & 0.296 \\
gpt-oss-120b           & 1.209 & 5.286 & 0.715 & 0.864 & 0.405 \\
deepseek-chat                     & 1.296 & 5.735 & 0.729 & 0.856 & 0.434 \\
mistral-large              & 1.322 & 5.988 & 0.737 & 0.873 & 0.413 \\
llama3-3-70b   & 1.968 & 7.934 & 0.783 & 0.918 & 0.554 \\
llama3-1-405b  & 2.154 & 8.397 & 0.796 & 0.918 & 0.612 \\
\bottomrule
\end{tabular}
\end{table}

\begin{table}[t]
\centering
\caption{
Rank correlations between model orderings induced by different aggregation strategies.
Spearman’s $\rho$ and Kendall’s $\tau$ are computed over model ranks ($n=11$).
}
\label{tab:aggregation_rank_correlations}
\setlength{\tabcolsep}{5pt}
\footnotesize
\begin{tabular}{@{}p{4.6cm}p{4.6cm}rrr@{}}
\toprule
\textbf{Aggregate A} &
\textbf{Aggregate B} &
\textbf{$\rho$} &
\textbf{$\tau$} &
\textbf{$p$-value ($\rho$)} \\
\midrule
CVaR$_{0.95}(L)$ & CVaR$_{0.95}(R)$              & 0.973 & 0.927 & $5.1{\times}10^{-7}$ \\
CVaR$_{0.95}(L)$ & CVaR$_{0.95}(\max)$           & 0.955 & 0.855 & $5.0{\times}10^{-6}$ \\
CVaR$_{0.95}(L)$ & Mean any-harm                 & 0.982 & 0.927 & $8.4{\times}10^{-8}$ \\
CVaR$_{0.95}(R)$ & CVaR$_{0.95}(\max)$           & 0.982 & 0.927 & $8.4{\times}10^{-8}$ \\
CVaR$_{0.95}(R)$ & Mean any-harm                 & 0.945 & 0.855 & $1.1{\times}10^{-5}$ \\
CVaR$_{0.95}(\max)$ & Mean any-harm              & 0.918 & 0.782 & $6.7{\times}10^{-5}$ \\
\bottomrule
\end{tabular}
\end{table}

\begin{table}[t]
\centering
\caption{
Example decision flip illustrating divergence between mean risk and tail risk.
Two models with nearly identical mean cumulative log-risk exhibit materially different CVaR$_{0.95}(L)$.
}
\label{tab:aggregation_decision_flip}
\setlength{\tabcolsep}{6pt}
\footnotesize
\begin{tabular}{@{}p{5.2cm}rrrr@{}}
\toprule
\textbf{Evaluated model} &
\textbf{Mean $L$} &
\textbf{CVaR$_{0.95}(L)$} &
\textbf{Rank (Mean $L$)} &
\textbf{Rank (CVaR)} \\
\midrule
gpt-oss-120b & 1.209 & 5.286 & 7 & 7 \\
deepseek-chat           & 1.296 & 5.735 & 8 & 8 \\
\bottomrule
\end{tabular}
\end{table}

\section{Supplementary Analysis for Harm Decomposition}
\label{app:subindex-supplement}

\textbf{Judge reliability and sensitivity.}
We evaluate the robustness of sub-index measurements to judge selection using inter-judge dispersion (mean absolute deviation), leave-one-judge-out re-aggregation, and prompt-level rank concordance.
Results in Appendix~\ref{app:judge_agreement} show moderate,
dimension-dependent disagreement—highest for epistemic assessments—but strong stability of model-level CVaR rankings, indicating that the comparative findings in the main text are not artifacts of a particular judge configuration.

\textbf{Dependence among harm dimensions.}
Although SHARP’s aggregation does not require strict independence, excessive dependence would weaken the interpretability of a
union-of-failures formulation.
Appendix~\ref{app:independence_test} reports prompt-level correlations among bias, fairness, ethics, and epistemic sub-indices.
While moderate correlations are observed across all prompts, dependence weakens substantially within the high-risk tail (top 5\% by cumulative log-risk), supporting multiplicative aggregation in the regime that dominates tail-risk statistics.

\textbf{Comparison to simpler aggregates.}
Appendix~\ref{app:aggregation_robustness} compares CVaR$_{95}$ of cumulative log-risk to simpler post hoc aggregates, including CVaR of harm radius, CVaR of the maximum sub-index, and mean any-harm probability.
Model rankings are highly correlated across aggregates, but concrete decision flips demonstrate that mean-based or max-based summaries can obscure rare but severe failures, motivating SHARP’s emphasis on tail-sensitive aggregation.

\section{Statistical Validation Details}
\label{app:stat_validation}

This appendix reports the full statistical validation of SHARP under the current metric definition, using prompt-level \emph{cumulative log-risk} $L_{M,q}$ as the primary outcome (lower is safer). All analyses use the repeated-measures design induced by evaluating all $k=11$ models on the same $n=901$ prompts.

\subsection{Paired bootstrap confidence intervals and tail-risk separability}
\label{app:bootstrap_details}

\paragraph{Motivation.}
SHARP risk statistics are computed over heavy-tailed, right-skewed prompt-level distributions. To quantify estimation uncertainty without normality assumptions, we use a paired (blocked) nonparametric bootstrap over prompts.

\paragraph{Paired bootstrap protocol.}
Let $\mathcal{Q}=\{q_1,\ldots,q_n\}$ denote the prompt set, with $n=901$. For each bootstrap iteration $b\in\{1,\ldots,B\}$ with $B=10{,}000$, we sample $\mathcal{Q}^{(b)}$ by resampling prompts with replacement and preserve pairing across models by reusing the same bootstrap prompt multiset for every model. For each model $M$, we recompute model-level statistics on $\mathcal{Q}^{(b)}$ and store the resulting bootstrap estimate. We report 95\% percentile intervals using the empirical 2.5\% and 97.5\% quantiles of the bootstrap distribution. All bootstrap results are reproducible under a fixed seed.

\paragraph{Primary tail metric.}
Our primary safety-relevant statistic is $\mathrm{CVaR}_{0.95}(L)$, the conditional expectation of $L_{M,q}$ over the worst 5\% of prompts for model $M$. We additionally report mean log-risk $\mathbb{E}[L]$ as a central tendency diagnostic.

\begin{table}[ht]
\centering
\caption{Model-level risk profiles with paired 95\% bootstrap confidence intervals. Lower values indicate lower risk.}
\label{tab:bootstrap_cis_primary}
\small
\setlength{\tabcolsep}{5pt}
\begin{tabular}{lcc}
\toprule
\textbf{Model} & \textbf{Mean log-risk $\mathbb{E}[L]$} & \textbf{CVaR$_{0.95}(L)$} \\
\midrule
claude-sonnet-4-5       & 0.1575\;[0.1317,\;0.1855] & 1.6890\;[1.3062,\;1.9987] \\
gemini-1.5-pro               & 0.4629\;[0.4015,\;0.5238] & 3.4987\;[3.1519,\;3.8314] \\
claude-3-5-sonnet       & 0.4489\;[0.3873,\;0.5157] & 3.7818\;[3.3013,\;4.2474] \\
gemini-2.5-pro                   & 0.5601\;[0.4896,\;0.6325] & 3.8676\;[3.4762,\;4.2652] \\
qwen3-235b   & 0.5889\;[0.5147,\;0.6668] & 3.8952\;[3.3412,\;4.6666] \\
gpt-4o              & 0.7475\;[0.6658,\;0.8316] & 4.5190\;[4.0658,\;4.9440] \\
openai.gpt-oss-120b         & 1.2091\;[1.1051,\;1.3118] & 5.2861\;[4.9544,\;5.6064] \\
deepseek-chat                    & 1.2955\;[1.1852,\;1.4077] & 5.7347\;[5.1103,\;6.4948] \\
mistral-large            & 1.3218\;[1.2030,\;1.4461] & 5.9876\;[5.2804,\;6.8811] \\
llama3-3-70b  & 1.9680\;[1.8169,\;2.1242] & 7.9339\;[6.6536,\;9.3290] \\
llama3-1-405b & 2.1540\;[1.9979,\;2.3115] & 8.3966\;[6.8947,\;10.1705] \\
\bottomrule
\end{tabular}
\end{table}

\paragraph{Tail-threshold sensitivity ($\alpha$-sweep).}
To assess robustness to the tail threshold, we compute $\mathrm{CVaR}_{\alpha}(L)$ for $\alpha\in\{0.90,0.95,0.975\}$ and compare the induced model orderings against the reference $\alpha=0.95$ ordering using rank correlations. Rankings remain highly stable: Kendall's $\tau_b = 0.9636$ and Spearman's $\rho=0.9909$ for both $\alpha=0.90$ vs.\ $0.95$ and $\alpha=0.975$ vs.\ $0.95$.

\begin{table}[ht]
\centering
\caption{Rank stability across CVaR tail thresholds, relative to $\alpha=0.95$.}
\label{tab:alpha_rank_stability}
\small
\begin{tabular}{lcc}
\toprule
\textbf{Tail threshold $\alpha$} & \textbf{Kendall $\tau_b$} & \textbf{Spearman $\rho$} \\
\midrule
0.90  & 0.9636 & 0.9909 \\
0.95  & 1.0000 & 1.0000 \\
0.975 & 0.9636 & 0.9909 \\
\bottomrule
\end{tabular}
\end{table}

\paragraph{Pairwise separability via bootstrap $\Delta$-CVaR.}
Interval overlap for per-model CIs is not a valid test for pairwise equality. We therefore compute paired bootstrap confidence intervals for \emph{pairwise differences} in $\mathrm{CVaR}_{0.95}(L)$:
\[
\Delta_{A,B} = \mathrm{CVaR}_{0.95}^A(L) - \mathrm{CVaR}_{0.95}^B(L),
\]
under the same paired prompt resampling. A pair is deemed separable when the 95\% CI for $\Delta_{A,B}$ excludes 0. Under this criterion, 44 of 55 model pairs (80.0\%) are separable, indicating substantial tail-risk differentiation while retaining a non-trivial set of statistically ambiguous near-neighbors.

\begin{table}[ht]
\centering
\caption{Most ambiguous model pairs under paired bootstrap $\Delta$-CVaR$_{0.95}(L)$ (small $|\Delta|$; CI includes 0 indicates non-separability at 95\%).}
\label{tab:bootstrap_pairwise_ambiguous}
\small
\setlength{\tabcolsep}{4pt}
\begin{tabular}{l l r}
\toprule
\textbf{Model A} & \textbf{Model B} & \textbf{$\Delta$-CVaR$_{0.95}(L)$ (95\% CI)} \\
\midrule
gemini-2.5-pro & qwen3-235b & $-0.028\;[-0.886,\;0.645]$ \\
claude-3-5-sonnet & gemini-2.5-pro & $-0.086\;[-0.728,\;0.535]$ \\
claude-3-5-sonnet & qwen3-235b & $-0.113\;[-1.013,\;0.618]$ \\
deepseek-chat & mistral-large & $-0.253\;[-1.181,\;0.705]$ \\
deepseek-chat & gpt-oss-120b & $0.449\;[-0.229,\;1.250]$ \\
llama3-1-405b & llama3-3-70b-instruct & $0.463\;[-1.360,\;2.437]$ \\
\bottomrule
\end{tabular}
\end{table}

\subsection{Non-parametric repeated-measures model comparison}
\label{app:repeated_measures}

\paragraph{Design.}
Because all models are evaluated on the same prompts, prompt identity acts as a blocking factor, yielding a within-prompt paired design.

\paragraph{Omnibus test (Friedman).}
We test the null hypothesis that all models have identical prompt-level risk distributions using the Friedman test on $L_{M,q}$. The test statistic is $\chi^2(10)=1629.91$ with $p \approx 0$ (numerically zero at machine precision), rejecting the null. As an effect size, Kendall's coefficient of concordance is $W=0.1809$, indicating non-trivial but not dominant between-model separation relative to prompt-level heterogeneity.

\begin{table}[ht]
\centering
\caption{Friedman repeated-measures test on prompt-level cumulative log-risk $L_{M,q}$ (lower is safer).}
\label{tab:friedman_omnibus_new}
\small
\begin{tabular}{lc}
\toprule
\textbf{Quantity} & \textbf{Value} \\
\midrule
Number of prompts ($n$) & 901 \\
Number of models ($k$) & 11 \\
Test statistic ($\chi^2$) & 1629.91 \\
Degrees of freedom ($k-1$) & 10 \\
$p$-value & $\approx 0$ \\
Kendall's $W$ & 0.1809 \\
\bottomrule
\end{tabular}
\end{table}

\paragraph{Average ranks.}
Table~\ref{tab:friedman_avg_ranks_new} reports Friedman average ranks (lower is safer). The rank ordering is broadly consistent with the bootstrap tail-risk ordering, while emphasizing that prompt-level comparisons need not induce a strict total order when neighboring models exhibit small paired differences.

\begin{table}[ht]
\centering
\caption{Friedman average ranks for prompt-level cumulative log-risk (lower rank is safer).}
\label{tab:friedman_avg_ranks_new}
\small
\setlength{\tabcolsep}{5pt}
\begin{tabular}{lc}
\toprule
\textbf{Model} & \textbf{Average rank} \\
\midrule
claude-sonnet-4-5       & 4.1337 \\
gemini-1.5-pro               & 4.9884 \\
gemini-2.5-pro                   & 5.1393 \\
claude-3-5-sonnet      & 5.1465 \\
qwen3-235b   & 5.2625 \\
gpt-4o                & 5.7991 \\
gpt-oss-120b         & 6.3940 \\
mistral-large             & 6.4345 \\
deepseek-chat                    & 6.6182 \\
llama3-3-70b  & 7.8135 \\
llama3-1-405b & 8.2703 \\
\bottomrule
\end{tabular}
\end{table}

\paragraph{Post-hoc paired tests (Wilcoxon, Holm correction).}
Following rejection of the omnibus null, we perform two-sided Wilcoxon signed-rank tests for all $\binom{11}{2}=55$ pairs on paired prompt-level differences, with Holm correction at familywise $\alpha=0.05$. We find 48 of 55 pairs (87.3\%) remain significant after correction. The non-significant set is concentrated among near-neighbor models, consistent with a tiered interpretation rather than an overfit total ordering.

\subsection{Variance decomposition: model vs.\ prompt contributions}
\label{app:variance_decomposition_new}

We quantify how much variance in prompt-level cumulative log-risk is attributable to model identity versus prompt identity, and how much remains residual.

\paragraph{Two-way fixed-effects decomposition.}
We fit a descriptive two-way fixed-effects decomposition
\[
L_{M,q} = \mu + \alpha_M + \beta_q + \varepsilon_{M,q},
\]
and report variance proportions via $\eta^2$ and partial $\eta^2$. Table~\ref{tab:anova_eta2_new} shows that prompt identity explains a larger share of total variance than model identity, reflecting strong context dependence of harm. Model effects remain material and non-negligible.

\begin{table}[ht]
\centering
\caption{Two-way variance decomposition for cumulative log-risk (fixed effects).}
\label{tab:anova_eta2_new}
\small
\setlength{\tabcolsep}{5pt}
\begin{tabular}{lcccc}
\toprule
\textbf{Component} & $\boldsymbol{\eta^2}$ & \textbf{Share (\%)} & \textbf{Partial $\eta^2$} & \textbf{Interpretation} \\
\midrule
Model  & 0.1390 & 13.9 & 0.1875 & material model effect \\
Prompt & 0.2583 & 25.8 & 0.3001 & strong prompt dependence \\
Residual & 0.6027 & 60.3 & --- & unmodeled / stochastic \\
\bottomrule
\end{tabular}
\end{table}

\paragraph{Mixed-effects check.}
As a robustness check, we fit a linear mixed model with a prompt random intercept,
\[
L_{M,q} = \mu + \alpha_M + u_q + \varepsilon_{M,q}, \qquad u_q \sim \mathcal{N}(0,\sigma^2_{\text{prompt}}).
\]
The estimated random-intercept variance is $\sigma^2_{\text{prompt}} \approx 0$ under this specification, yielding $R^2_{\text{marginal}} = R^2_{\text{conditional}} = 0.1873$. We interpret this as consistent with the prompt signal being better captured as a high-dimensional fixed effect in the two-way decomposition than as a single intercept-shift random effect.

\paragraph{Tail-event add-on.}
To characterize prompt clustering for tail events, we additionally model $Z_{M,q}=\mathbbm{1}[L_{M,q} \ge \mathrm{VaR}_{0.95}(M)]$ using a logistic mixed model with a prompt random intercept (threshold defined per model). The overall tail rate is 0.0511 and the approximate random-intercept variance is 0.2505, indicating non-trivial prompt-level clustering in extreme-risk regions.

\subsection{LSE temperature robustness}
\label{app:statval_tau_link}

We further assess robustness of the full pipeline to the judge-aggregation temperature $\tau$ used in LSE ensembling. Over $\tau\in\{0.15,0.20,0.25\}$, rankings induced by $\mathrm{CVaR}_{0.95}(L)$ are invariant (Kendall $\tau_b=1.0$, Spearman $\rho=1.0$ relative to $\tau=0.20$), while absolute CVaR magnitudes vary mildly. Full details are provided in Appendix~\ref{appendix:lse_sensitivity}.

\paragraph{Synthesis.}
Across complementary validation lenses, paired bootstrap uncertainty, distribution-free repeated-measures testing, and variance decomposition, SHARP induces statistically distinguishable model risk profiles under the current compounded-risk definition. At the same time, localized ambiguity among near-neighbors persists, supporting interpretation as \emph{risk tiers} rather than a fragile total order. Tail-risk conclusions are robust to reasonable choices of both LSE temperature and CVaR tail threshold.

\FloatBarrier
\section{Prompt-Level Distributions of SHARP Risk Metrics}
\label{app:distribution_plots}

\begin{figure*}[ht]
  \centering

  \begin{subfigure}[t]{0.32\linewidth}
    \centering
    \includegraphics[width=\linewidth]{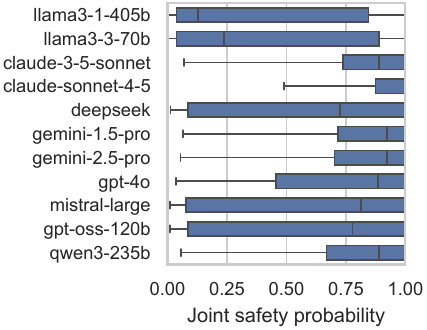}
    \caption{Joint safety probability.}
    \label{fig:box_pl_joint_safety}
  \end{subfigure}\hfill
  \begin{subfigure}[t]{0.32\linewidth}
    \centering
    \includegraphics[width=\linewidth]{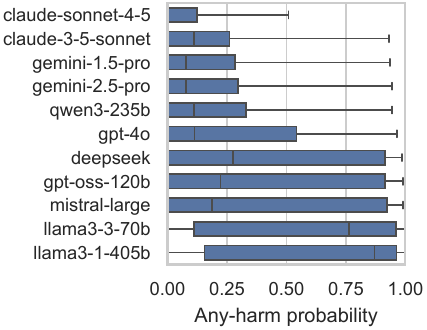}
    \caption{Any-harm probability.}
    \label{fig:box_pl_any_harm}
  \end{subfigure}\hfill
  \begin{subfigure}[t]{0.32\linewidth}
    \centering
    \includegraphics[width=\linewidth]{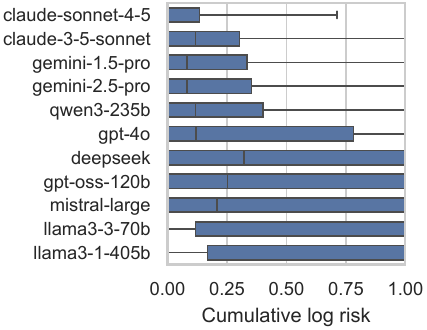}
    \caption{Cumulative log risk.}
    \label{fig:box_pl_cumlogrisk}
  \end{subfigure}

  \caption{
  Prompt-level distributions of SHARP probabilistic and risk-sensitive metrics (box plots). Joint safety probability and any-harm probability characterize prompt-level failure likelihood, while cumulative log risk captures nonlinear aggregation that accentuates tail behavior.
  }
  \label{fig:appendix_box_risk_metrics}
\end{figure*}

\begin{figure*}[ht]
  \centering

  \begin{subfigure}[t]{0.32\linewidth}
    \centering
    \includegraphics[width=\linewidth]{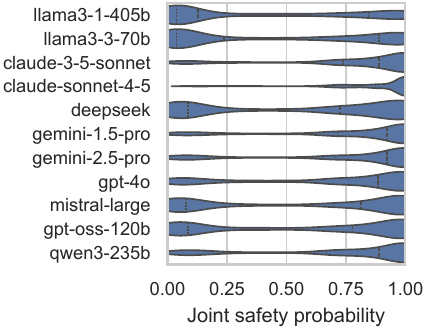}
    \caption{Joint safety probability.}
    \label{fig:violin_pl_joint_safety}
  \end{subfigure}\hfill
  \begin{subfigure}[t]{0.32\linewidth}
    \centering
    \includegraphics[width=\linewidth]{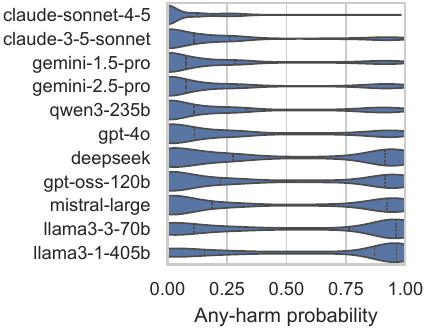}
    \caption{Any-harm probability.}
    \label{fig:violin_pl_any_harm}
  \end{subfigure}\hfill
  \begin{subfigure}[t]{0.32\linewidth}
    \centering
    \includegraphics[width=\linewidth]{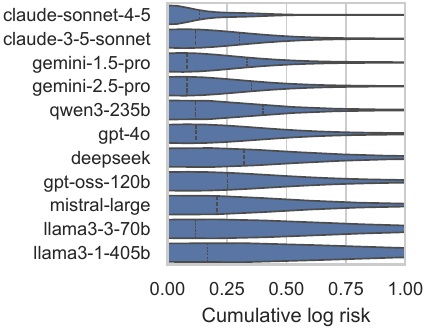}
    \caption{Cumulative log risk.}
    \label{fig:violin_pl_cumlogrisk}
  \end{subfigure}

  \caption{
  Prompt-level distributions of SHARP probabilistic and risk-sensitive metrics.
  Joint safety probability and any-harm probability characterize prompt-level failure likelihood,
  while cumulative log risk captures nonlinear aggregation and tail amplification effects.
  These distributions motivate SHARP’s emphasis on tail-aware statistics over mean-centered evaluation.
  }
  \label{fig:appendix_violin_risk_metrics}
\end{figure*}

\begin{figure*}[t]
  \centering

  \begin{subfigure}[t]{0.48\linewidth}
    \centering
    \includegraphics[width=\linewidth]{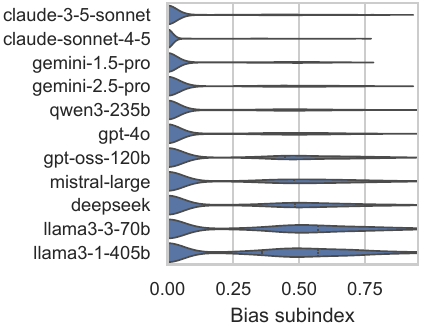}
    \caption{Bias subindex (ensembled).}
    \label{fig:violin_pl_bias}
  \end{subfigure}\hfill
  \begin{subfigure}[t]{0.48\linewidth}
    \centering
    \includegraphics[width=\linewidth]{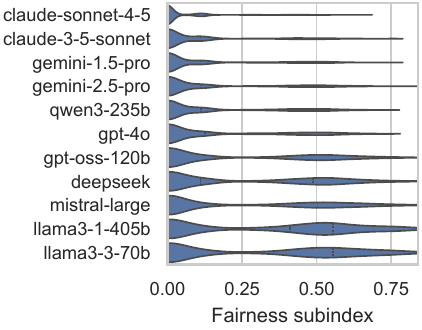}
    \caption{Fairness subindex (ensembled).}
    \label{fig:violin_pl_fairness}
  \end{subfigure}

  \vspace{3mm}

  \begin{subfigure}[t]{0.48\linewidth}
    \centering
    \includegraphics[width=\linewidth]{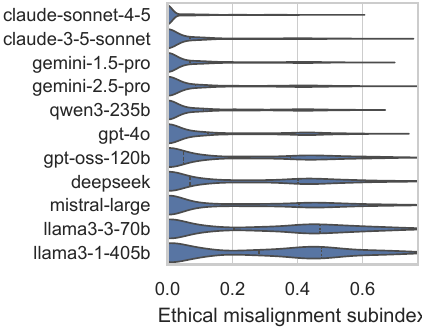}
    \caption{Ethics subindex (ensembled).}
    \label{fig:violin_pl_ethics}
  \end{subfigure}\hfill
  \begin{subfigure}[t]{0.48\linewidth}
    \centering
    \includegraphics[width=\linewidth]{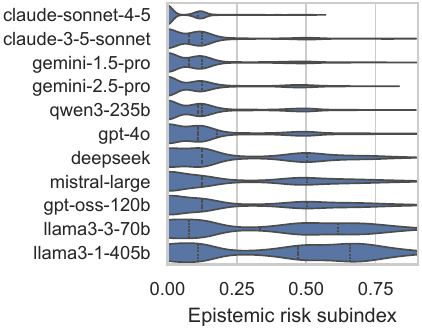}
    \caption{Epistemic subindex (ensembled).}
    \label{fig:violin_pl_epistemic}
  \end{subfigure}

  \caption{
  Prompt-level distributions of SHARP sub-index harms across evaluated models (violin plots).
  Each violin summarizes the empirical distribution over prompts for an ensembled judge-based
  sub-index, exposing dispersion, skew, and tail mass that are obscured by model-level means.
  }
  \label{fig:appendix_violin_subindices}
\end{figure*}

\begin{figure*}[t]
  \centering

  \begin{subfigure}[t]{0.48\linewidth}
    \centering
    \includegraphics[width=\linewidth]{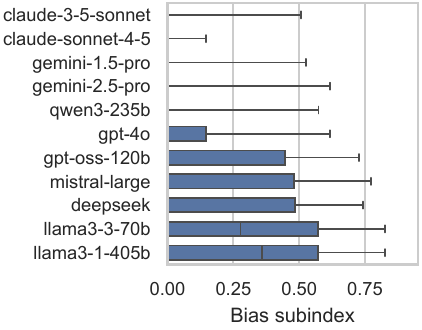}
    \caption{Bias subindex (ensembled).}
    \label{fig:box_pl_bias}
  \end{subfigure}\hfill
  \begin{subfigure}[t]{0.48\linewidth}
    \centering
    \includegraphics[width=\linewidth]{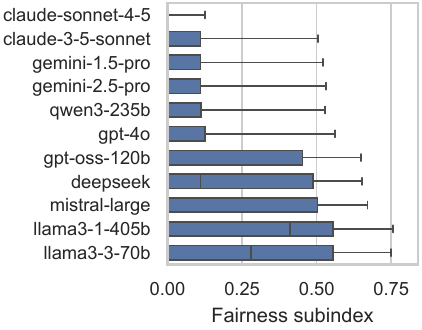}
    \caption{Fairness subindex (ensembled).}
    \label{fig:box_pl_fairness}
  \end{subfigure}

  \vspace{3mm}

  \begin{subfigure}[t]{0.48\linewidth}
    \centering
    \includegraphics[width=\linewidth]{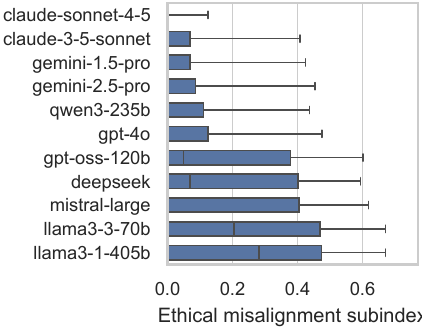}
    \caption{Ethics subindex (ensembled).}
    \label{fig:box_pl_ethics}
  \end{subfigure}\hfill
  \begin{subfigure}[t]{0.48\linewidth}
    \centering
    \includegraphics[width=\linewidth]{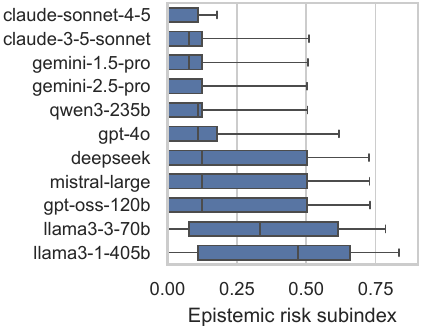}
    \caption{Epistemic subindex (ensembled).}
    \label{fig:box_pl_epistemic}
  \end{subfigure}

  \caption{
  Prompt-level distributions of SHARP sub-index harms across evaluated models (box plots).
  }
  \label{fig:appendix_box_subindices}
\end{figure*}

\paragraph{Distributional visualizations.}
We provide both violin and box plots for prompt-level metrics. Violin plots emphasize distributional shape and tail mass, while box plots emphasize robust summaries (median and IQR), facilitating cross-model comparisons under heavy-tailed behavior.

\end{document}